\documentclass[11pt,a4paper]{article}
\usepackage{lineno}

\usepackage{times,latexsym}
\usepackage{url}
\usepackage[T1]{fontenc}

\usepackage[acceptedWithA]{tacl2021v1}

\usepackage{xspace,mfirstuc,tabulary}

\newif\iftaclinstructions
\taclinstructionsfalse  
\iftaclinstructions
\newcommand{\instr}
\fi

\iftaclpubformat

\else

\fi

\DeclareUnicodeCharacter{266B}{}
\usepackage[most]{tcolorbox}

\usepackage{graphicx}

\usepackage{float}
\usepackage{multirow}
\usepackage{graphicx}
\usepackage{listings}
\usepackage{array}
\usepackage{tabularx}
\usepackage{booktabs}
\usepackage{array}
\usepackage{adjustbox}
\usepackage{tikz}
\usepackage{xcolor}
\usepackage{varwidth}
\usepackage{arydshln}
\usepackage{pgfplots}
\pgfplotsset{compat=1.18}
\usepackage{graphicx}
\usepackage{svg}
\usepackage{bm}
\usepackage{inconsolata}
\usepackage{enumitem}
\usepackage{subcaption}
\usepackage{hyperref}
\usepackage[dvipsnames,table]{xcolor}

\title{No Shortcuts to Culture: Indonesian Multi-hop Question Answering for Complex Cultural Understanding}

\author{
  Vynska Amalia Permadi$^{1,2}$ \quad  Xingwei Tan$^{1}$ \quad
  Nafise Sadat Moosavi$^{1}$ \quad
  Nikos Aletras$^{1}$
  \\
  $^{1}$School of Computer Science, University of Sheffield, United Kingdom \\
  $^{2}$Department of Informatics, Universitas Pembangunan Nasional ``Veteran'' Yogyakarta, Indonesia
  \\
  {\texttt{\{vpermadi1,xingwei.tan,n.s.moosavi,n.aletras\}@sheffield.ac.uk}}
}

\date{}

\begin{document}
\maketitle
\newcommand{\hficon}{\raisebox{-0.15em}{\includegraphics[height=1.3em]{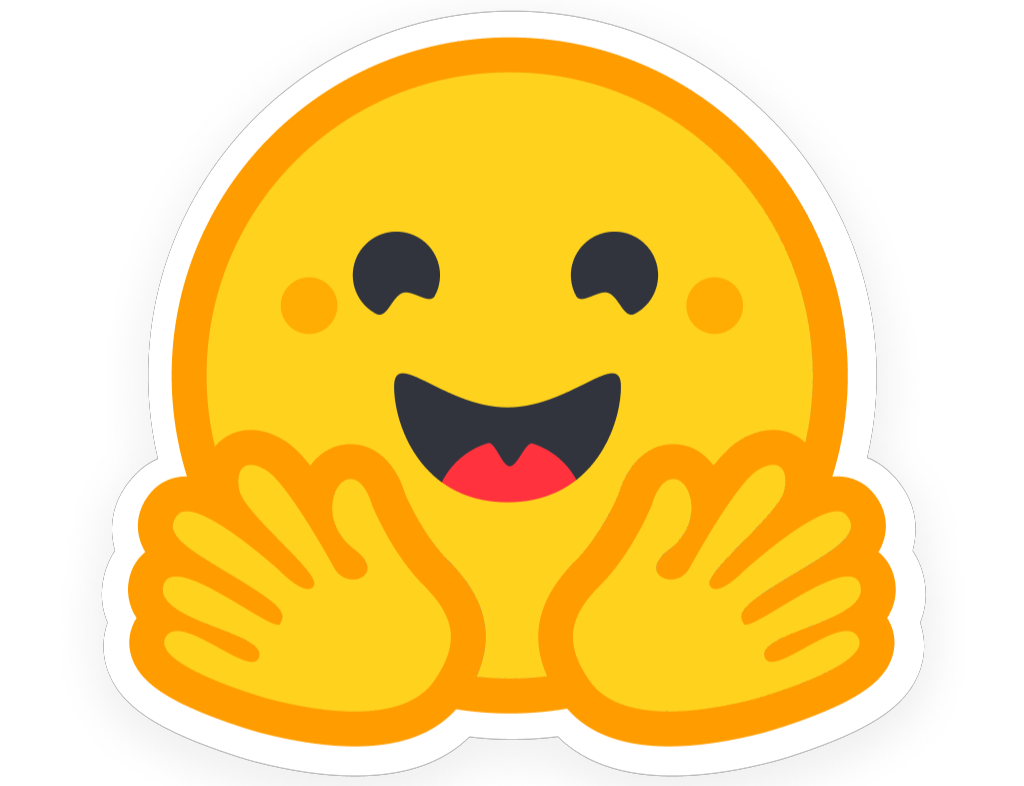}}}

\begin{abstract}
Understanding culture requires reasoning across context, tradition, and implicit social knowledge, far beyond recalling isolated facts. Yet most culturally focused question answering (QA) benchmarks rely on single-hop questions, which may allow models to exploit shallow cues rather than demonstrate genuine cultural reasoning. In this work, we introduce ID-MoCQA, the first large-scale multi-hop QA dataset for assessing the cultural understanding of large language models (LLMs), grounded in Indonesian traditions and available in both English and Indonesian. We present a new framework that systematically transforms single-hop cultural questions into multi-hop reasoning chains spanning six clue types (e.g., commonsense, temporal, geographical). Our multi-stage validation pipeline, combining expert review and LLM-as-a-judge filtering, ensures high-quality question-answer pairs. Our evaluation across state-of-the-art models reveals substantial gaps in cultural reasoning, particularly in tasks requiring nuanced inference. ID-MoCQA provides a challenging and essential benchmark for advancing the cultural competency of LLMs.\footnote{Dataset is available at \hficon\url{https://huggingface.co/datasets/vynsk/ID-MoCQA}}
\end{abstract}

\section{Introduction}
Developing large language models (LLMs) that can truly understand unwritten social norms, diverse local traditions, and cultural knowledge is important for the development of systems that can effectively avoid cultural insensitivities or misunderstandings, reinforcing stereotypes, or causing offence \cite{NEURIPS2024_BLEnD,10.1162/COLI.a.14}.

\begin{figure}[t]
    \centering
    \includegraphics[width=0.9\linewidth]{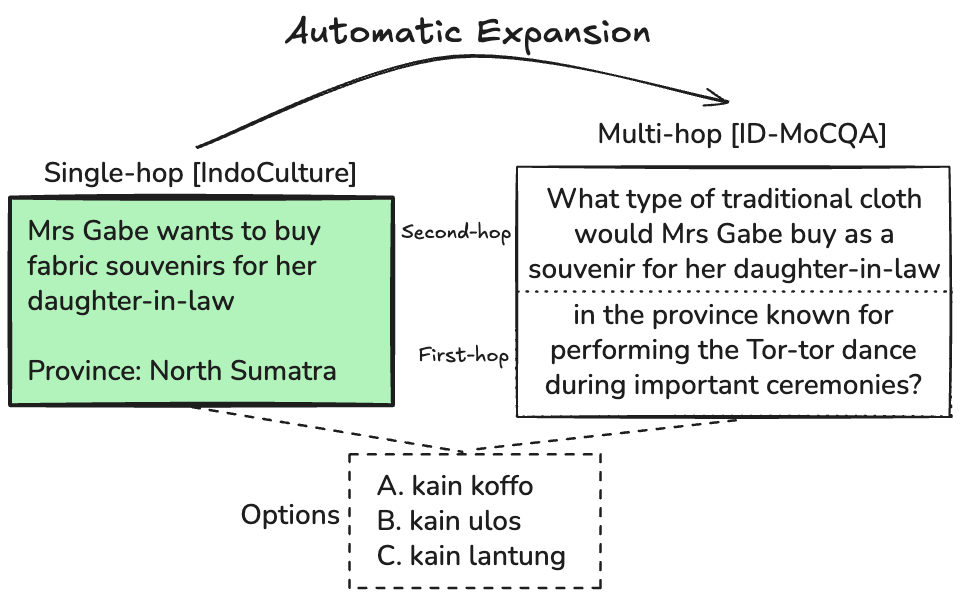}
    \caption{Single to multi-hop transformation from IndoCulture \citep{koto-etal-2024-indoculture} to ID-MoCQA. Left: Original question about fabric souvenirs with origin province. Right: Our expansion requires first predicting the province (\textit{North Sumatra}) through cultural clues (\textit{Tor-tor dance}), then answering the question.}
    \label{fig:example_introduction}
\end{figure}

Recent research has focused on developing resources for assessing cultural knowledge of LLMs, especially on low-resource languages~\cite{NEURIPS2024_BLEnD,putri-etal-2024-llm}. However, the majority of these benchmarks are built around single-hop question answering, where the answer can be retrieved directly from a single fact or cue. While effective for measuring factual knowledge, such setups often fail to probe whether models can reason through more complex, interrelated cultural knowledge concepts \cite{wang2024kulturebench,kim-etal-2024-click,koto-etal-2024-indoculture}. 

By contrast, multi-hop QA aims to evaluate deeper reasoning. Datasets such as HotpotQA \citep{yang-etal-2018-hotpotqa}, 2WikiMultiHopQA \citep{ho-etal-2020-constructing}, and MuSiQue \citep{trivedi-etal-2022-musique} challenge models to combine multiple pieces of evidence to reach an answer, reducing the likelihood of shortcut exploitation. Applying this multi-hop paradigm to the cultural domain is a natural next step, one that enables us to test whether models can interpret cultural clues, connect context, and infer appropriate practices.

In this work, we present a new framework to address this gap by transforming culturally grounded single-hop questions into two-hop QA instances that simulate realistic cultural reasoning (Figure \ref{fig:example_introduction}). Our multi-hop structure tests whether LLMs understand not just cultural facts, but their contextual application: models must first identify relevant cultural context, then select practices appropriate to that context. To ensure correctness, we first prompt an LLM to add one intermediate reasoning step that connects to the original context while ensuring that the added step is relevant for answering the final multi-hop question. We further incorporate a multi-stage validation process that combines expert annotation and LLM-as-a-judge \cite{Zheng23LLM-as-a-judge} evaluation. This framework results in ID-MoCQA, the first large-scale multi-hop cultural QA dataset focused on a single national context: Indonesia. ID-MoCQA contains 15,590 questions, equally distributed across six clue types and two languages (English and Indonesian). Our extensive evaluation across a range of open, frontier, and region-specific LLMs reveals
clear limitations in multi-hop cultural reasoning, even among top-performing models. Our contributions can be summarized as follows:

\begin{itemize}
    \item We propose a new comprehensive framework for generating cultural multi-hop questions from existing single-hop data.
    \item We release \textbf{ID-MoCQA}, a human-verified dataset of 15,590 multi-hop questions in Indonesian and English about Indonesian culture generated by our proposed framework.
    \item We conduct extensive evaluation, including a diverse collection of open and frontier LLMs on \textbf{ID-MoCQA}, revealing persistent challenges in cultural multi-hop reasoning and establishing the dataset as a robust benchmark for future research.
\end{itemize}

\begin{figure*}[!t]
  \centering
  \includegraphics[width=0.9\textwidth]{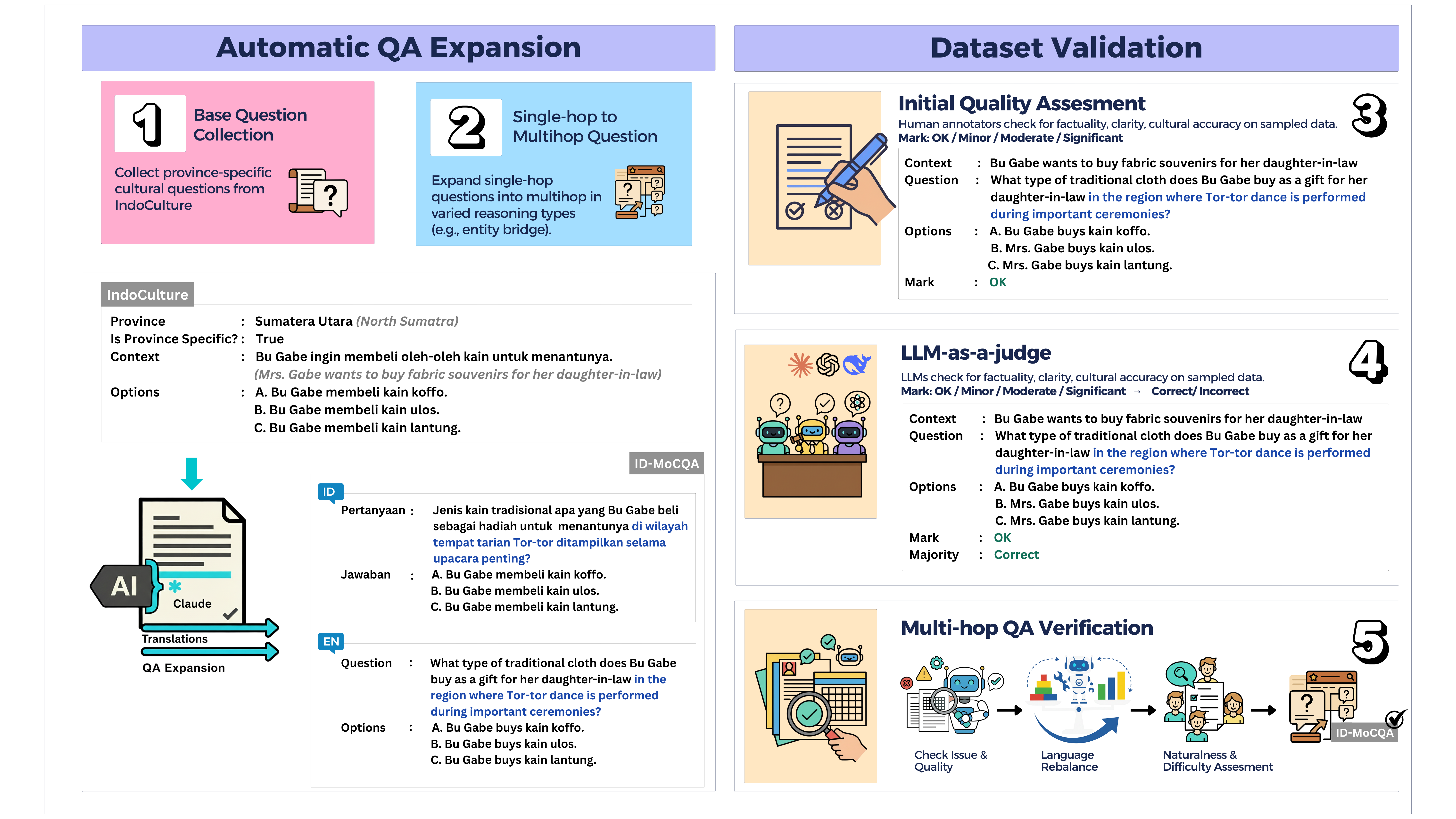}
  \caption{ID-MoCQA dataset creation pipeline. \textbf{Left (Automatic QA Expansion)}: (1) Collection of province-specific questions from IndoCulture; (2) Expansion to multi-hop questions using \textit{Claude-3.7-Sonnet} with varied clue types, generating bilingual (Indonesian/English) versions. \textbf{Right (Dataset Validation)}: (3) Human assessment of factuality, clarity, and cultural accuracy; (4) Quality verification via LLM-as-a-judge; (5) Multi-hop verification to check quality, ensure language balance, and assess naturalness and difficulty.}
  \label{fig:pipeline_overview}
\end{figure*}

\section{Related Work}
\subsection{Cultural Competence in Sociolinguistics}
Cultural competence is the ability to communicate and act appropriately within different communities and contexts, including knowing when, how, and to whom communications are suitable \citep{Hymes1972,Byram1997}. The same statement can be acceptable in one community but not in another, depending on factors such as social relationships, status, and context \citep{Goffman1967,BrownLevinson1987}. Previous work has identified systematic patterns in cultural differences. Analysing high- and low-context communication shows whether meaning relies on shared cultural knowledge or on explicit verbal content \citep{Hall1976}. Cultural dimensions, including individualism-collectivism and power distance, also describe how cultures differ in communication expectations \citep{Hofstede2011}. These patterns may also guide behaviour in common social situations such as greetings, requests, and expressions of gratitude \citep{Wierzbicka1994}.

\subsection{Cultural QA Benchmarks}
Measuring and modeling culture in LLMs has emerged as a critical research area \citep{adilazuarda-etal-2024-towards}, with growing recognition that cultural understanding encompasses more than factual knowledge \citep{zhou-etal-2025-culture}. In response, specialized cultural QA benchmarks have been developed to evaluate LLM performance across diverse socio-cultural contexts. BLEnD \citep{NEURIPS2024_BLEnD} offers over 52,000 QA pairs on daily life and socio-cultural topics, spanning 16 countries and 13 languages. NativQA \citep{hasan2024nativqa} provides a semi-automatic framework for building culturally aligned QA datasets. It includes approximately 64,000 manually annotated and 55,000 automatically generated pairs across seven languages and 18 topics from nine regions. Complementing these approaches, WorldValuesBench (HRMCR) \citep{zhao-etal-2024-worldvaluesbench} evaluates multicultural value understanding through scenarios inspired by the World Values Survey, while CultureAtlas \citep{fung2024CultureAtlasmassivelymulticultural} compiles culturally rich Wikipedia knowledge representing diverse sub-country regions and ethnolinguistic groups. INCLUDE \citep{romanou2024includeevaluatingmultilinguallanguage} addresses multilingual evaluation gaps with over 197,000 QA pairs from local exams in 44 languages, grounding evaluation in regional settings rather than English translations.

Recognizing the need for deeper cultural authenticity and representation, researchers have developed language-specific benchmarks that capture specific regional contexts and linguistic nuances in under-represented languages. For Indonesian, IndoCloze \citep{koto-etal-2022-cloze} offers short narratives assessing story comprehension and commonsense reasoning with causal and temporal understanding requirements. ID-CSQA \citep{putri-etal-2024-llm} includes 9,000 culturally relevant commonsense QA pairs for Indonesian and Sundanese, created through LLMs and human annotation. COPAL-ID \citep{wibowo-etal-2024-copal}, crafted by native speakers, focuses on natural causal reasoning in both standard and Jakartan Indonesian to capture local context. IndoCulture \citep{koto-etal-2024-indoculture} is a benchmark developed through collaborative discussions with Indonesian natives, ensuring comprehensive coverage of diverse cultural aspects from 11 provinces across 6 islands of the Indonesian archipelago. Each province represents distinct ethnic groups, regional languages, and religious practices. Korean benchmarks include KULTURE Bench \citep{wang2024kulturebench}, featuring cultural news, idioms, and poetry, and CLIcK \citep{kim-etal-2024-click}, offering 1,995 QA pairs from official exams and textbooks with fine-grained cultural and linguistic knowledge annotations. CAMeL \citep{naous-etal-2024-beer} provides an Arabic benchmark contrasting Arab and Western cultures across tasks such as story generation and sentiment analysis. More recently, CulturalBench \citep{chiu-etal-2025-culturalbench} introduced 1,696 human-verified questions covering 45 global regions through a human-AI collaborative approach, which represents an advance in robust cultural knowledge evaluation. Despite these advances, all existing cultural QA datasets focus exclusively on single-hop questions.

\subsection{Automatic QA Data Construction}
LLMs have demonstrated potential for QA dataset generation through prompting strategies. CulturePark \citep{NEURIPS2024_77f089cd} uses LLMs to generate diverse single-hop cross-cultural reasoning questions at scale. \citet{shah-etal-2024-improving} introduce planned query guidance using few-shot examples to enable systematic multi-hop reasoning over knowledge graphs. Cultural applications include ID-CSQA for Indonesian commonsense reasoning \cite{putri-etal-2024-llm}, NativQA for multilingual cultural alignment \cite{hasan2024nativqa}, and WikiQA-IS for Icelandic cultural knowledge \cite{thorunn-arnardottir-etal-2025-wikiqa}. However, the intersection of cultural authenticity and multi-hop complexity presents unique challenges. Ensuring both cultural accuracy and valid reasoning structures requires careful methodology, particularly when dealing with culture-specific knowledge that may be underrepresented in LLM training data.

\section{Multi-hop QA Generation Framework}
Our aim is to expand single-hop cultural questions to multi-hop.
Figure~\ref{fig:pipeline_overview} presents our comprehensive framework for building ID-MoCQA, which consists of two main components: (1) \textit{Automatic QA Expansion} (Steps 1--2), which systematically transforms single-hop questions into multi-hop questions through LLM-guided generation; and (2) \textit{Dataset Validation} (Steps 3--5), which implements a multi-stage quality assurance process combining human expertise with LLM verification to ensure dataset reliability.

\begin{table*}[t]
\renewcommand{\arraystretch}{1.2}
\tiny
\setlength{\tabcolsep}{3pt}
\centering
\begin{tabularx}{\textwidth}{p{1.2cm} p{7cm} X}
\hline
\textbf{Type} & \textbf{Key Rule / Focus} & \textbf{Example} \\
\hline
Entity & Use only exact entity names (people, historical figures, cultural artifacts) uniquely associated with one province. Avoid geographical features and compound entities with ``and''. & What musical instrument is central to wedding ceremonies \textcolor{blue}{in the province where Cut Nyak Dhien led resistance against Dutch colonialism?} \\
Geographical & Use only geographical features located exclusively in target province. Never use cross-boundary features (rivers, mountain ranges, watersheds). & What traditional fabric is produced  \textcolor{blue}{in the region where the Derawan Islands are located?}\\
Temporal & Use significant events with specific dates/time periods that uniquely identify exactly one province. Verify historical accuracy of all temporal references. & What traditional dance is performed during harvest festivals  \textcolor{blue}{in the province where the Majapahit Empire had its capital from 1293 to 1527?} \\
Commonsense & Begin with ``If...'' scenario (max 2 sentences) using descriptive language to uniquely identify province through distinctive cultural attributes without naming popular items. &  \textcolor{blue}{If A is female and inherits her mother's property in a western Indonesian province with matrilineal traditions based on Adat Perpatih customary law}, what traditional dish would she serve to welcome visitors? \\
Comparison & Use single, precise comparison (rankings, superlatives, relative metrics) with verifiable data that identifies exactly one province. Specify year for population data. & What traditional ceremony is practised  \textcolor{blue}{in the Indonesian province with the third highest number of UNESCO World Heritage Sites?} \\
Intersection & First statement [S1] identifies multiple provinces (3-5) using general categories. Second statement [S2] narrows to exactly province. Both conditions must be distinct attributes. & What special dish is prepared during Eid celebrations  \textcolor{blue}{in the Indonesian province with BOTH active volcanoes AND the largest Buddhist temple in the world?}\\
\hline
\end{tabularx}
\caption{Examples of multi-hop prompt types andtheir key principles. Blue text represents the first-hop clues that suggest the provinces, and the black text represents the original IndoCulture question.}
\label{tab:multi-hop-prompt-types}
\end{table*}

\subsection{Base Single-hop Question Collection} 
First, we derive our initial single-hop QA pairs from the IndoCulture dataset (Figure~\ref{fig:pipeline_overview}, Step 1). Each  pair has a label distinguishing between province-specific and general cultural elements (\textit{True} or \textit{False}), indicating whether the cultural element uniquely pertains to the province. Figure~\ref{fig:example_introduction} (top left) shows an example of a single-hop instance with its associated location. We exclusively select instances marked as \textit{True}, representing practices unique to particular provinces, so we can use the province names as a first-hop link (i.e., \textit{region where Tor-tor dance is performed during important ceremonies}). This yields 1{,}847 province-specific QA pairs, which serve as the foundation for multi-hop expansion.

\subsection{From Single-hop to Multi-hop}
Our framework transforms the manually curated high-quality single-hop questions from IndoCulture (Figure~\ref{fig:pipeline_overview}, Step 2). We build the first-hop by converting the province information into clues that require geographical, temporal, commonsense, or other cultural reasoning. This design introduces multi-step reasoning into the question while preserving the cultural authenticity of IndoCulture. In IndoCulture, the input consists of the province name, context (i.e., \textit{Mrs. Gabe wants to buy fabric souvenirs for her daughter-in-law}), and options (i.e., \textit{A. Mrs Gabe buys kain koffo; B. Mrs. Gabe buys kain ulos; C. Mrs. Gabe buys kain lantung.}). To create more challenging questions that test multi-step cultural reasoning, our expansion process consists of the following steps: (1) first-hop link type creation; and (2) bilingual multi-hop question generation, where we simultaneously generate culturally authentic questions in both Indonesian and English using an LLM. Figure~\ref{fig:pipeline_overview} (bottom left) shows our overall process for expanding single-hop questions to multi-hop. 

\paragraph{First-hop question type (clue type).} Following \citet{Mavi2024MHQAtype}, we design six types of cultural clues: commonsense, comparison, entity, geographical, intersection, and temporal. For each type, we develop specific transformation guidelines and create distinct prompts with tailored instructions and few-shot examples (Appendix~\ref{app:multi-hop}). These cultural clues are automatically generated by prompting LLMs to produce reasoning-based multi-hop questions. To answer the transformed question, models must first determine which province the cultural clues refer to. Table~\ref{tab:multi-hop-prompt-types} summarises the key principles and provides examples for each question type. For example, the \textsc{entity} clue type uses prompts designed to identify provinces through specific cultural items such as the \textit{Tor-tor dance}, which is a unique traditional dance from \textit{North Sumatra}. \textit{The province serves as a first-hop entity while the original IndoCulture context becomes the second-hop cultural question}.

\paragraph{Bilingual multi-hop question generation.} To enable broader accessibility to our data, we perform multi-hop question generation through two sequential sub-processes for each clue type, simultaneously generating culturally authentic questions in both Indonesian and English:

\begin{itemize}
    \item \textbf{Statement to question conversion:} We convert each original context statement into a question while removing direct province mentions (if any). For example, given the example in Figure~\ref{fig:pipeline_overview}, this is transformed to \textit{What type of traditional cloth does Bu Gabe buy as a gift for her daughter-in-law}.
    \item \textbf{First-hop link type integration:} We add first-hop clues based on the selected clue type that require reasoning to identify the target province. These use only indirect cultural references without mentioning provinces, cities, or regencies directly. The final result combines both steps: \textit{What type of traditional cloth does Bu Gabe buy as a gift for her daughter-in-law in the region where Tor-tor dance is performed during important ceremonies?}
\end{itemize}

We generate questions in Indonesian and English using \textit{Claude-3.7-Sonnet} \cite{claude3.7}  with temperature $=1$.  We translate the text into the target languages while keeping the culture-specific terms (e.g., \emph{Rumoh Aceh}) unchanged. The prompt template is shown in Appendix~\ref{app:multi-hopPrompt}. We apply this process to 1{,}847 IndoCulture questions across six clue types in both languages, yielding 22{,}164 instances.

\section{Dataset Validation}
\label{sec:validation}

\subsection{Initial Quality Assessment}
\label{subsec:initial-verification}
As in the first validation stage (Figure~\ref{fig:pipeline_overview}, Step 3), we first conduct manual verification on 3,000 randomly sampled instances (in both languages) from our dataset. We reviewed each question and classified them into four quality categories.
\textbf{OK:} Questions have no substantive issues, or only negligible stylistic flaws (e.g., occasional repeated terms, inconsistencies in punctuation, or missing people's names) that do not affect meaning or clarity.
\textbf{Minor:} Questions contain slightly unnatural yet understandable translations, including suggestions for improvements in format, tone, or clarity.
\textbf{Moderate:} Questions where cultural clues are ambiguous and could apply to multiple provinces, making the intended answer uncertain.
\textbf{Significant:} Questions include factually incorrect statements, leak the correct answer, or are incomprehensible.

\begin{table}
  \centering
  \scriptsize
  \begin{tabular}{lcc}
    \hline
    \textbf{Mark} & \textbf{Count} & \textbf{Percentage \%} \\
    \hline
    \textbf{OK} & $\textbf{1,712}$ & $\textbf{57.07\%}$ \\
    Minor & $242$ & $8.07\%$ \\
    Moderate & $260$ & $8.67\%$ \\
    Significant & $786$ & $26.20\%$ \\
    \midrule
    Total & $3,000$ &  $100.00\%$ \\
    \hline
  \end{tabular}

  \caption{Distribution of quality marks from manual verification of 3,000 randomly sampled multi-hop instances.}
  \label{tab:Datadist}
\end{table}
We found that 57.07\% of the sampled questions meet acceptable standards (OK), 26.20\% contain significant errors. 
Analysis by clue type shows that \textsc{Intersection} and \textsc{Comparison} questions demonstrate higher rates of ``significant'' issues (46.8\% and 68.0\%, respectively), indicating that \textit{Claude-3.7-Sonnet} struggles more with generating these question types. \textsc{Comparison} questions show particularly low quality rates, with only 25.4\% marked as OK. Common issues include incorrect factual statements in \textsc{Comparison} criteria, e.g., a question about \textit{the province with the third largest area of wetland rice cultivation in Kalimantan} incorrectly refers to South Kalimantan (second in agricultural land among Kalimantan provinces according to 2024 data).

\subsection{LLM-as-a-Judge}
\label{subsec:LLM-as-a-Judge}
While manual verification provides insights into data quality, evaluating manually all 22,164 instances is not feasible. Hence,  we implement an LLM-as-a-judge \cite{Zheng23LLM-as-a-judge} using three frontier models (Figure~\ref{fig:pipeline_overview}, step 4): \textit{GPT-4o} \cite{gpt4o}, \textit{Claude-3.7-Sonnet}, and \textit{DeepSeek-V3} \cite{deepseekai2024deepseekv3technicalreport}. The evaluation assesses key aspects of question quality including factual accuracy, structural coherence, and linguistic quality (guidelines in Appendix~\ref{app:evaluation_prompt}).

\paragraph{Multi-Annotator Validation.} To assess the reliability of the LLM-as-a-Judge, we conduct a second validation round with two independent annotators (native Indonesian speakers who have lived in multiple Indonesian provinces) on the same  subset as in \hyperref[subsec:initial-verification]{§4.1}. The annotators validated the questions using the same  scale and guidelines as the LLM-as-a-judge. When they disagree, a third annotator provides the final judgment.

\begin{table}[t]
  \centering
  \scriptsize
  \begin{tabular}{lc}
    \hline
    \textbf{Question Type} & \textbf{Cohen's $\kappa$} \\
    \hline
    Geographical & 0.75 \\
    Entity & 0.68 \\
    Temporal & 0.55 \\
    Commonsense & 0.48 \\
    Comparison & 0.42 \\
    Intersection & 0.35 \\
    \midrule
    \textbf{Average} & \textbf{0.54} \\
    \hline
  \end{tabular}
  \caption{Inter-annotator agreement (Cohen's $\kappa$) across question types on sample questions.}
  \label{tab:kappa}
\end{table}

\paragraph{Inter-annotator Agreement.} Table~\ref{tab:kappa} presents Cohen's $\kappa$ for each question type. The average $\kappa$ across question types is $0.54$, with individual types ranging from $0.35$ to $0.75$. These scores indicate fair to moderate agreement \citep{Artstein2008}. \textsc{Geographical} questions have the highest agreement score, while \textsc{Intersection} questions the lowest. This variation indicates that cultural reasoning complexity varies by category, with location-specific and entity-identification tasks proving easier for consistent annotation than \textsc{Intersection} or \textsc{Comparison} tasks.
We analyse the disagreements between human annotators (see Appendix~\ref{app:human_disagreements}). We find that the disagreements often reflect difference perspectives in judgment rather than annotation errors. As \citet{fleisig-etal-2023-majority} note, when annotators disagree on subjective judgments, this often reflects genuine differences in interpretation.

\paragraph{Validating LLM-as-a-Judge with human annotations.} We convert human and LLM ratings into binary labels: \textit{OK} and \textit{Minor} to \textit{Acceptable}, \textit{Moderate} and \textit{Significant} to \textit{Unacceptable}. First, we evaluate the LLM judge's decisions against the human gold standard on the dual-annotated instances. The automated filtering achieves a precision of 0.78 and a recall of 0.82. This indicates that while the LLM judge identifies most acceptable questions (high recall), approximately 22\% of accepted instances may contain false positives and false negatives.
We calculate Intraclass Correlation Coefficient (ICC) to quantify the consistency of observations within groups. We use a two-way random effects model for absolute agreement. As shown in Table~\ref{tab:icc-by-class}, the LLM judge achieves moderate agreement with human annotators, with higher agreement on acceptable than unacceptable instances. This pattern, combined with the precision and recall scores, shows that the LLM judge effectively identifies high-quality questions but shows more variation when detecting problematic instances.

\begin{table}[t]
\centering
\scriptsize
\begin{tabular}{lccc}
\toprule
\textbf{Quality Class} & \textbf{\shortstack{LLM vs\\Ann.1}} & \textbf{\shortstack{LLM vs\\Ann.2}}  \\
\midrule
Acceptable     & 0.75 & 0.72  \\
Unacceptable   & 0.66 & 0.63  \\
\midrule
Overall        & 0.71 & 0.68  \\
\bottomrule
\end{tabular}
\caption{Intraclass Correlation Coefficient (ICC) by quality class.}
\label{tab:icc-by-class}
\end{table}

While automated quality assessment has limitations, the LLM-as-a-Judge provides a practical solution for filtering large-scale generated data when comprehensive manual annotation is infeasible. Based on the validation results, we apply the following filtering rules: instances receiving majority votes of \textit{Acceptable} from at least two of the three LLMs are retained, while any instance marked as \textit{Significant} by any single LLM is automatically rejected. This process filtered the dataset to 12,939 instances in both Indonesian and English. Given the observed false positive rate, we implement additional structure-based verification in the next validation stage to further validate the annotations.

\subsection{Question Structure Verification}
As part of the final validation stage (Figure~\ref{fig:pipeline_overview}, Step 5), we implement a two-stage verification process (Appendix~\ref{app:verification_prompts}).
\paragraph{Phase 1: Issue Detection.} The first phase uses \textit{Claude-3.7-Sonnet} to identify and correct two specific structural issues. We ask \textit{Claude-3.7-Sonnet} to assess whether the multi-hop question contains phrases directly copied from the provided answer options. If it does, it rewrites the copied text. To detect invalid province name (\texttt{contains\_province}), we ask \textit{Claude-3.7-Sonnet} to examine whether province names appear as geographical location references (e.g., ``from Bali''). If it does, we replace them with indirect references while preserving cultural terminology (e.g., ``Batik Aceh'').

\paragraph{Phase 2: Quality Assessment.} 
For the questions verified by the previous phase, \textit{Claude-3.7-Sonnet} assesses whether they meet multi-hop requirements based on a two-step reasoning structure, sequential logic, and cultural question alignment. \textit{Claude-3.7-Sonnet} also suggests which type of revision is needed. If a question needs only minor improvements (i.e., grammar, clarity, formatting, capitalization), it will receive automated refinements.If a question needs a fundamental restructuring of the reasoning flow, it will be flagged as ``[NEEDS MAJOR REVISION]'' and removed from the dataset.
Less than 1\% of the questions are removed due to their failure to meet multi-hop requirements in the final assessment.
Manual verification of the ``[NEEDS MAJOR REVISION]'' questions confirm that they are not the desired multi-hop questions. 

\subsection{Question Language Rebalance}
Continuing Step 5 of our framework (Figure~\ref{fig:pipeline_overview}), we identify questions that only exist in one language based on the ID-type pairs. If an ID-type pair appears only in English, we ask \textit{Claude-3.7-Sonnet} to translate the question into Indonesian, and vice versa for questions in Indonesian. We ensure the translation preserves cultural terms, proper nouns, and traditional item names. 

\subsection{Naturalness and Difficulty Assessment}
\label{subsec:naturalness}
In the final part of Step 5 (Figure~\ref{fig:pipeline_overview}), we perform an evaluation of linguistic naturalness and cognitive difficulty across all questions. Three native Indonesian speakers independently assess the questions following specific guidelines (Appendix~\ref{app:annotation-guide}). 

\paragraph{Naturalness.} Annotators rated both Indonesian and English versions on a three-point scale: \textit{Natural}, \textit{Acceptable}, or \textit{Unnatural}. Using majority voting of their ratings, we identify questions that need revision. About 8\% of Indonesian questions rated \textit{Unnatural} due to translation errors or incorrect subjects/names, while 3\% were \textit{Acceptable} due to minor grammar or translation issues.
For English questions, about 7\% are flagged as \textit{Unnatural} and 2\% as \textit{Acceptable}. All questions rated as \textit{Unnatural}  are manually revised.

\paragraph{Cognitive Difficulty.} On average, 44.8\% of the questions were rated as \textit{Hard}, 25.9\% as \textit{Moderate}, and 29.2\% as \textit{Easy}. This highlights the challenging nature of our dataset.

\begin{table}[!t]
\centering
\scriptsize
\begin{tabular*}{0.9\columnwidth}{@{\extracolsep{\fill}}lc lc@{}}
\toprule
\textbf{Topic} & \textbf{\#Q} & \textbf{Province} & \textbf{\#Q} \\
\midrule
Food & $1{,}335$ & West Sumatra & $1{,}072$ \\
Wedding & $1{,}175$ & Papua & $891$ \\
Art & $1{,}025$ & North Sumatra & $850$ \\
Family relations & $695$ & Aceh & $808$ \\
Pregnancy \& kids & $667$ & South Kalimantan & $783$ \\
Socio-religious & $646$ & Bali & $747$ \\
Religious holiday & $497$ & South Sulawesi & $714$ \\
Death & $431$ & Central Java & $653$ \\
Daily activities & $384$ & West Java & $582$ \\
Traditional games & $330$ & East Java & $469$ \\
Fisheries \& trade & $311$ & East Nusa Tenggara & $226$ \\
Agriculture & $299$ &  &  \\
\bottomrule
\end{tabular*}
\caption{Questions across topics and provinces.}
\label{tab:Distributions}
\end{table}

\subsection{Final Dataset}
\label{subsec:final-dataset}
Following our verification framework in \hyperref[subsec:initial-verification]{§4.1} to \hyperref[subsec:naturalness]{§4.5}, ID-MoCQA contains 15,590 instances evenly distributed across Indonesian and English with 7,795 instances per language. The distribution of question types is uneven due to the difficulties in generating high-quality questions for different categories, as shown in Table \ref{tab:type_lexical_stats}. \textsc{Comparison} questions have the lowest verification success rate, representing the smallest category with only 730 instances per language. These questions require generating superlatives or performing differential analysis between cultural elements, but are frequently marked as ``Significant'' during LLM-as-a-judge evaluation due to factual inaccuracies. The generated questions often contain incorrect ranking statements or unverifiable comparative assertions about cultural practices across provinces.

Each question requires sequential reasoning: first identifying the target province through cultural clues, then answering the province-specific cultural question. Questions span 11 Indonesian provinces across 6 islands, covering 12  cultural topics. Table \ref{tab:Distributions} presents the distribution across topics and provinces.

\paragraph{Semantic and Lexical Analysis.}
We conducted automated linguistic analysis using GPT-4o-mini to extract part-of-speech tags, named entities, temporal expressions, and Indonesian cultural terms from all 7,795 English questions (Appendix~\ref{app:semantic}). Analysis was performed on the English version to ensure consistent part-of-speech extraction, as Indonesian cultural terms are preserved identically across both language versions. Questions in ID-MoCQA average 24.3 words, with 12,381 adjectives (1.59 per question), 54,297 nouns (6.97 per question), and 14,967 verbs (1.92 per question). \textsc{COMMONSENSE} questions average 30.9 words with 2.5 adjectives per question, supporting their conditional scenario structures, while \textsc{GEOGRAPHICAL} questions average 18.9 words with 0.8 adjectives per question, reflecting more direct reference patterns. The data includes 1,274 unique person names and 1,447 unique location names. \textsc{ENTITY} questions reference 496 unique persons, while \textsc{TEMPORAL} questions cite 280 unique locations. ID-MoCQA contains 2,398 unique Indonesian cultural terms appearing 8,499 times (1.09 per question), preserved in their original language across both versions. \textsc{COMMONSENSE} and \textsc{INTERSECTION} questions contain 680 and 633 unique terms respectively, while \textsc{GEOGRAPHICAL} contains 294. Approximately 38.1\% of questions (2,972) include temporal expressions spanning historical periods, contemporary events, and cultural calendars. Table ~\ref{tab:type_lexical_stats} presents complete statistics across question types.

\begin{table}[!t]
\centering
\resizebox{\columnwidth}{!}{
\begin{tabular}{lccccccccc}
\toprule
\textbf{Type} & \textbf{\#QA} & \textbf{Avg.} & \textbf{Adj.} & \textbf{N.} & \textbf{V.} &
\multicolumn{3}{c}{\textbf{Unique Entities}} \\ 
 &  & \textbf{Culture} & \textbf{/Q} & \textbf{/Q} & \textbf{/Q} &
\textbf{Pers.} & \textbf{Loc.} & \textbf{ID} \\
\midrule
Commonsense  & $1{,}424$ & 30.9 & 2.5 & 8.7 & 2.4 & 391 & 223 & 680 \\
Comparison   &   $730$   & 21.5 & 1.6 & 5.8 & 1.7 & 171 &  96 & 231 \\
Entity       & $1{,}447$ & 20.6 & 0.9 & 5.9 & 1.6 & 496 & 246 & 440 \\
Geographical & $1{,}508$ & 18.9 & 0.8 & 5.4 & 1.5 & 327 & 274 & 294 \\
Intersection & $1{,}279$ & 27.2 & 2.2 & 7.8 & 2.1 & 369 & 234 & 633 \\
Temporal     & $1{,}407$ & 26.1 & 1.6 & 7.4 & 2.0 & 377 & 280 & 419 \\
\midrule
\textbf{Overall} & \textbf{$7{,}795$} & \textbf{24.3} & \textbf{1.59} & \textbf{6.97} & \textbf{1.92} & 
\textbf{1{,}274} & \textbf{1{,}447} & \textbf{2{,}398} \\
\bottomrule
\end{tabular}
} 
\caption{Overview of lexical and cultural characteristics across question types. The table shows the number of QA pairs, the average cultural length (Avg.~Culture), and the average number of adjectives, nouns, and verbs per question. The final columns list the counts, including persons (Pers.), locations (Loc.), and identification terms (ID) in each question type. 
}
\vspace{-8pt}

\label{tab:type_lexical_stats}
\end{table}

\section{Experimental Setup}
\subsection{Models}
\label{sec:tested_llm}

\paragraph{Frontier LLMs:} \textit{GPT-5} \cite{gpt5}, \textit{DeepSeek-V3}, and \textit{Claude-3.7-Sonnet}. 
    
\paragraph{Multilingual open models:} \textit{Gemma2-27B-Instruct} \citep{gemma_2024}, \textit{Llama3.3-70B-Instruct} \citep{Llama3.3}, \textit{Llama3.1-8B} \citep{Llama3.1}, \textit{Qwen2.5-72B-Instruct} and \textit{Qwen2.5-7B} \citep{qwen2.5}.
    
\paragraph{Region-specific open models:} \textit{Merak-7B} \citep{ichsan2023merak} and \textit{SeaLLM-7B} \citep{nguyen-etal-2024-seallms} are trained on Indonesian and are the top two performing models on the IndoCulture. This inclusion allows us to assess whether regional specialisation provide advantages for the multi-hop reasoning in ID-MoCQA.

Given a multi-hop cultural question, a model needs to first identify the relevant Indonesian province based on the clues, then answer a province-specific cultural question about that region. Each question provides three options, which are from the original IndoCulture dataset for the final answer and requires open-ended text generation for province identification. All experiments are conducted using prompts designed to obtain both province identification and final answer selection (Appendix~\ref{app:eval_prompts}).

\subsection{Human Baseline}
We recruit three university graduates (different to the original annotators), who are native Indonesian speakers, to answer all 7,795 ID-MoCQA questions. The guidelines are shown in Appendix~\ref{app:annotation-guide}. The participants need to identify the target province first, then select one of the three options without access to external tools.

\renewcommand{\arraystretch}{1.2}
\begin{table*}[t]
\centering
\tiny
\resizebox{0.8\textwidth}{!}{
\begin{tabular}{clccccccc}
\toprule
 & \textbf{Model} & \textbf{Comm.} & \textbf{Comp.} & \textbf{Entity} & \textbf{Geo.} & \textbf{Inter.} & \textbf{Tempo.} & {\textbf{Overall}} \\
\midrule

\multirow{11}{*}{EN} 
& Claude-3.7-Sonnet   & $\mathbf{81.32}$ & $79.86$ & $\mathbf{80.51}$ & $\mathbf{81.76}$ & $81.24$ & $\mathbf{81.59}$ & $\mathbf{81.15}$ \\
& DeepSeek-V3         & $76.19$ & $73.01$ & $75.33$ & $76.66$ & $77.56$ & $74.84$ & $75.81$ \\
& GPT-5              & $79.85$ & $\mathbf{82.60}$ & $79.82$ & $80.90 $ & $\mathbf{81.31}$ & $80.95$ & $80.74$ \\
\cmidrule{2-9}
& Llama3.3-70B-IT     & $69.73$ & $66.58$ & $68.00$ & $69.10$ & $67.94$ & $69.44$ & $68.65$ \\
& Qwen2.5-72B-IT      & $67.49$ & $66.44$ & $67.31$ & $66.25$ & $67.08$ & $66.95$ & $66.95$ \\
& Gemma2-27B-IT       & $65.87$ & $65.21$ & $66.76$ & $68.37$ & $67.32$ & $64.61$ & $66.47$ \\
& Llama3.1-8B         & $53.93$ & $54.52$ & $55.15$ & $55.84$ & $54.42$ & $53.16$ & $54.52$ \\
& Qwen2.5-7B          & $54.56$ & $55.07$ & $54.94$ & $57.16$ & $52.46$ & $53.73$ & $54.69$ \\
& Merak-7B           & $52.81$ & $52.60$ & $54.73$ & $54.38$ & $50.82$ & $53.16$ & $53.19$ \\
& SeaLLM-7B          & $51.47$ & $51.23$ & $51.14$ & $52.79$ & $51.21$ & $51.60$ & $51.62$ \\
\midrule
\midrule

\multirow{10}{*}{ID} 
& Claude-3.7-Sonnet   & $\mathbf{82.23}$ & $80.55$ & $\mathbf{80.72}$ & $\mathbf{83.16}$ & $\mathbf{82.17}$ & $82.30$ & $\mathbf{81.98}$ \\
& DeepSeek-V3         & $77.46$ & $75.89$ & $75.88$ & $76.66$ & $77.72$ & $77.04$ & $76.83$ \\
& GPT-5              & $81.18$ & $\mathbf{82.47}$ & $79.89$ & $81.43$ & $81.24$ & $\mathbf{82.59}$ & $81.37$ \\
\cmidrule{2-9}
& Llama3.3-70B-IT     & $72.96$ & $69.73$ & $70.97$ & $72.75$ & $70.91$ & $70.65$ & $71.49$ \\
& Qwen2.5-72B-IT      & $70.01$ & $68.77$ & $69.59$ & $69.69$ & $69.59$ & $69.23$ & $69.54$ \\
& Gemma2-27B-IT       & $67.49$ & $64.25$ & $68.35$ & $68.63$ & $68.73$ & $66.17$ & $67.53$ \\
& Llama3.1-8B         & $57.58$ & $54.93$ & $57.50$ & $59.48$ & $58.33$ & $56.43$ & $57.60$ \\
& Qwen2.5-7B          & $54.14$ & $53.15$ & $54.39$ & $54.38$ & $51.84$ & $56.43$ & $54.18$ \\
& Merak-7B           & $48.42$ & $51.10$ & $52.66$ & $52.19$ & $50.20$ & $52.24$ & $51.14$ \\
& SeaLLM-7B          & $49.58$ & $51.51$ & $51.07$ & $52.45$ & $50.12$ & $51.17$ & $50.97$ \\
\cmidrule{2-9}
& Human (n=3) & $69.85{\pm}4.6$ & $70.05{\pm}3.9$ & $69.89{\pm}4.4$ & $70.53{\pm}4.2$ & $70.45{\pm}4.7$ & $69.23{\pm}6.0$ & $69.99{\pm}4.6$ \\

\bottomrule
\end{tabular}
}
\caption{Accuracy (\%) across multi-hop clue types and overall in English and Indonesian.}
\label{tab:accuracy_MHQA}
\end{table*}

\section{Results and Analysis}

\subsection{Human Performance}
\label{subsec:human-performance}
Humans achieve 70.0\% multi-hop accuracy, with individual performance ranging from 66.6\% to 75.3\%. First-hop accuracy is 95.1\%. The 25.1\% gap between first-hop and multi-hop shows that identifying the location is much easier.

\paragraph{Difficulty ratings and performance.} The difficulty assessments align with their performance. On average, 44.8\% of questions were rated as Hard, corresponding to the 30\% failure rate in multi-hop accuracy. Individual difficulty perceptions varied considerably, with ratings ranging from 32.3\% to 53.1\% for Hard questions. The native speaker who rated the most questions as Hard achieved 68.1\% multi-hop accuracy, while the one who rated the fewest as Hard achieved 75.3\% multi-hop accuracy.

\subsection{Zero-shot Results}
\begin{figure*}[!t]
    \centering
    \begin{subfigure}[t]{0.49\textwidth}
        \centering
        \includegraphics[width=\textwidth]{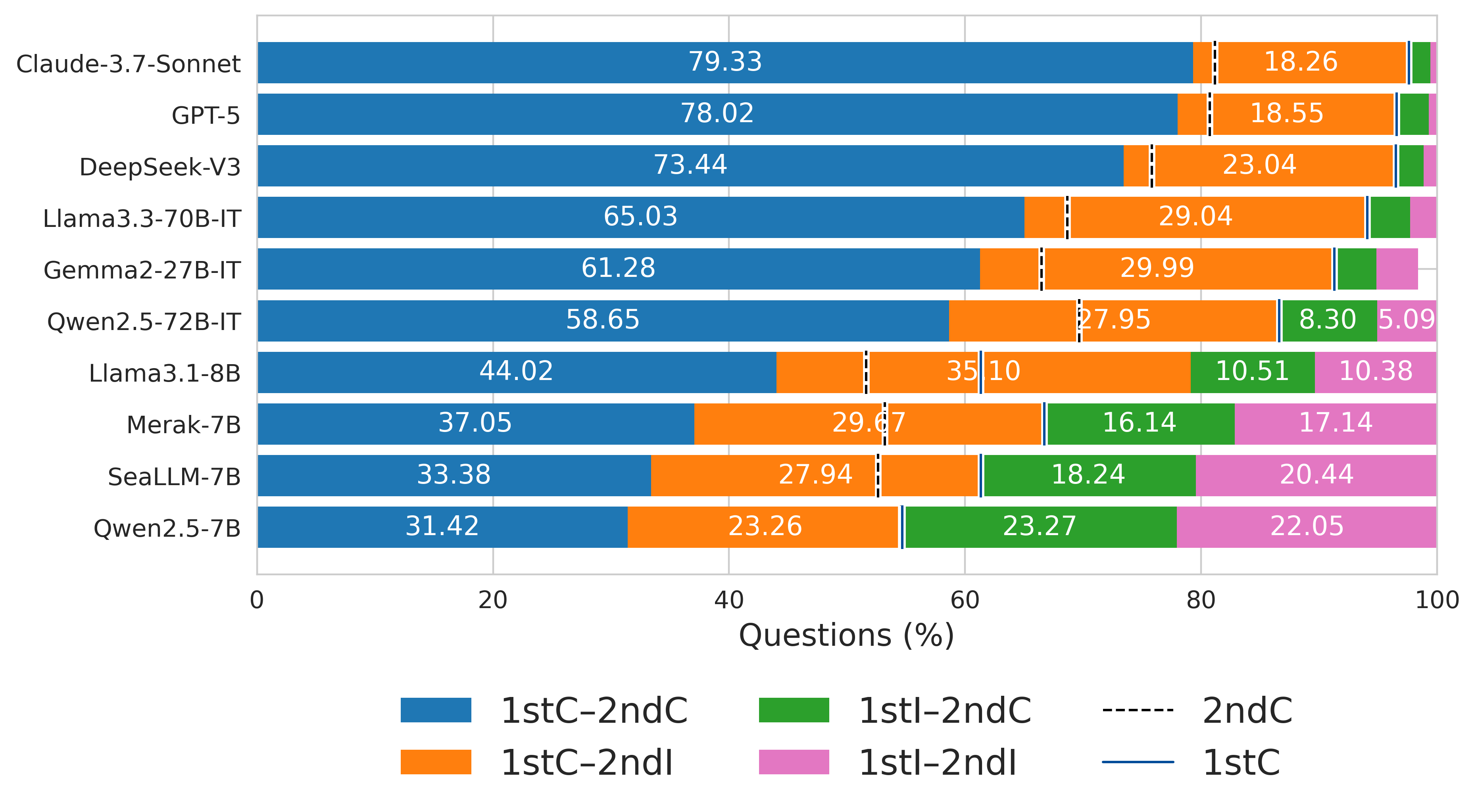}
        \caption{English}
        \label{fig:accuracy_breakdown_en}
    \end{subfigure}
    \hfill
    \begin{subfigure}[t]{0.49\textwidth}
        \centering
        \includegraphics[width=\textwidth]{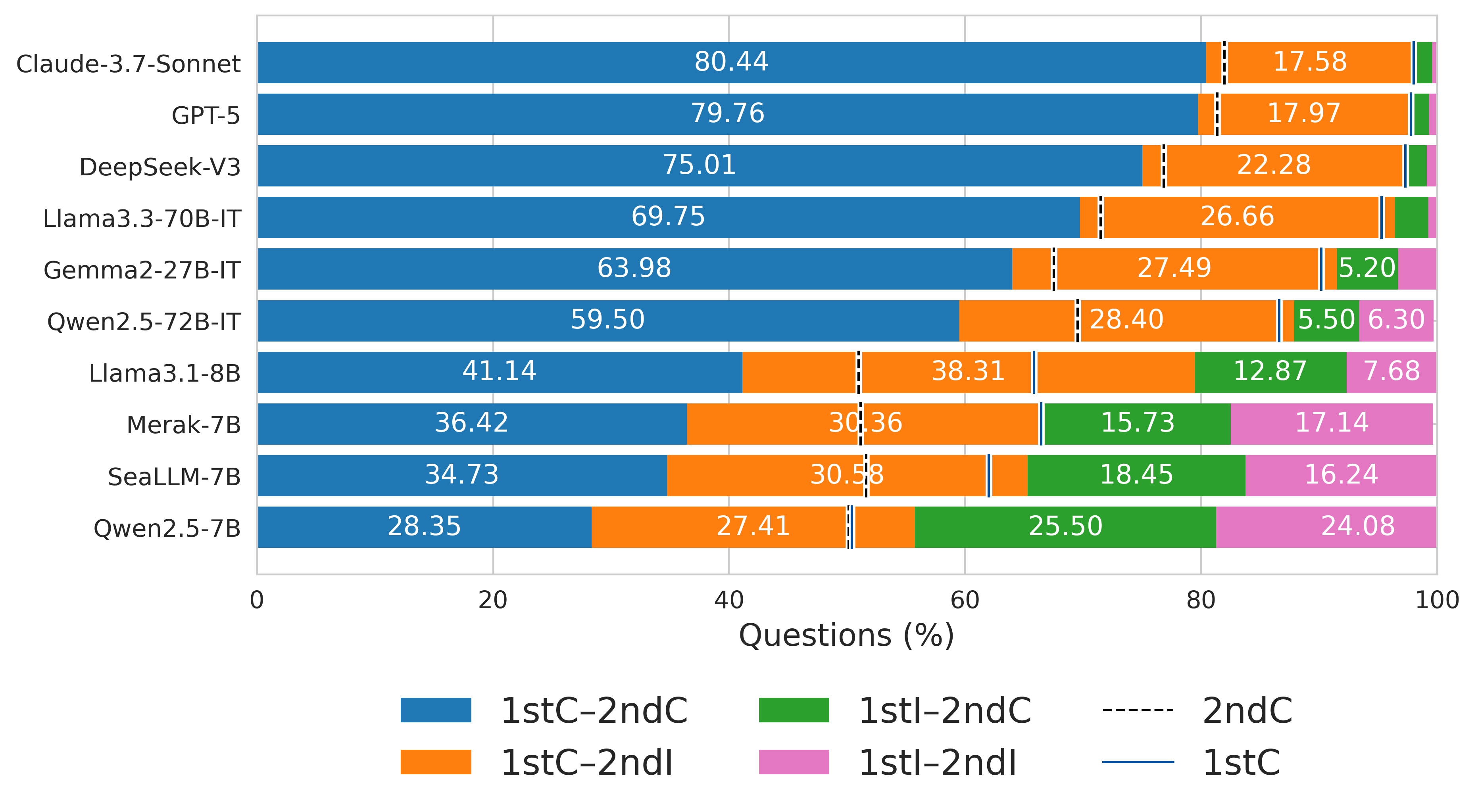}
        \caption{Indonesian}
        \label{fig:accuracy_breakdown_id}
    \end{subfigure}
    \caption{Breakdown of model predictions (\%) by first-hop (province-level) and second-hop (final answer) correctness for English and Indonesian. 
    } 
\vspace{-8pt}
    \label{tab:firsthop_secondhop_breakdown}
\end{figure*}

\paragraph{Frontier LLMs outperform humans.} Table \ref{tab:accuracy_MHQA} shows \textit{GPT-5 }and \textit{Claude-3.7-Sonnet} achieve the highest multi-hop accuracy in both languages, surpassing the human baseline by over 10\% in Indonesian. \textit{DeepSeek-V3} follows closely, also outperforms humans. 

\paragraph{Geographic knowledge differences explain the gap.} Frontier LLMs outperform humans across all 11 provinces, but the gap varies depending on the status of those provinces. Bali, with its distinct Hindu culture and prominent role in the tourism industry, along with West Java and Central Java on Java island, which is the most populous island and location of the capital, are more familiar to most Indonesians. On these provinces, humans score 84\% on average while models score 86\%. However, when the questions are about provinces (e.g., Papua and Aceh) that are away from the economic and political centers, human performance drops to 65\% while frontier LLMs maintain 77\%. This pattern suggests that LLMs' training data provides more balanced coverage of regional cultural knowledge than individuals' lived experience.\footnote{To verify this gap reflects genuine knowledge differences rather than dataset biases, we analyze answer position distribution and question length effects. Human errors spread evenly across answer options with no position bias, and question length shows no correlation with the gap.}

\paragraph{Larger models perform better than smaller models in Indonesian.} \textit{GPT-5} and \textit{Claude-3.7-Sonnet} achieve the highest performance, and they both perform better in Indonesian than English. \textit{DeepSeek-V3} and other larger models follow the same pattern. 
However, smaller models show inconsistent language preferences. \textit{Merak-7B} and \textit{SeaLLM-7B} perform worse in Indonesian in most clue types despite being fine-tuned on Indonesian Wikipedia. \textit{Qwen2.5-7B} shows a similar trend with slightly lower performance in Indonesian. In contrast, \textit{Llama3.1-8B} achieves approximately 3 percentage points higher accuracy in Indonesian (57.60 vs. 54.52). Merak and SeaLLM achieve $\sim$53\% accuracy in IndoCulture's single-hop questions, but drop to 51.14\% and 50.97\% respectively in ID-MoCQA's multi-hop questions. This shows that although specialized models can handle single-hop cultural questions effectively, the additional complexity of multi-hop reasoning poses challenges to smaller models, even in their target languages. The performance gap between larger and smaller models increases for complex reasoning tasks, regardless of language specialization.

\paragraph{Performance on clue types varies by both model type and language.} Table \ref{tab:accuracy_MHQA} reveals that no single clue type is universally easier or harder for all models. Each model shows a distinct profile of strengths and weaknesses. For instance, while \textsc{Comparison} represents the most challenging type for \textit{Claude-3.7-Sonnet}, \textit{DeepSeek-V3}, and \textit{Llama3.3-70B-IT}, \textit{GPT-5} achieves their peak performance on this type in English. Smaller models show more diverse patterns: \textit{Qwen2.5-7B} peaks on \textsc{Geographical}, while \textit{Merak-7B} peaks on \textsc{Entity}. Similarly, \textit{Merak-7B} struggles most on \textsc{Intersection} in English, yet \textit{DeepSeek-V3} shows its highest accuracy on this type. In contrast, \textit{Llama3.1-8B} struggles most with \textsc{Temporal}, while \textit{SeaLLM-7B} peaks on \textsc{Geographical}. This divergence indicates that different models have developed distinct reasoning capabilities.

The pattern shifts in Indonesian. \textit{Merak-7B}'s lowest-performing type shifts from \textsc{Intersection} in English to \textsc{Commonsense} in Indonesian, dropping below 49\% accuracy. \textit{SeaLLM-7B} shifts from \textsc{Entity} to \textsc{Commonsense} as its weakest type. \textit{GPT-5} maintains \textsc{Entity} as its weakest type in both languages, while \textit{Qwen2.5-72B-IT} shifts from \textsc{Geographical} in English to \textsc{Temporal} in Indonesian. Their peak performance types also change: \textit{GPT-5} moves from \textsc{Comparison} to \textsc{Temporal} as its strongest type, and \textit{Gemma2-27B-IT} shifts from \textsc{Geographical} to \textsc{Intersection}. 

\paragraph{Frontier LLMs excel at province prediction but not so good at selecting final answers.} Figure \ref{tab:firsthop_secondhop_breakdown} shows frontier models achieve over 96\% first-hop accuracy but are 18\% to 23\% lower when considering both steps.
Correct first-hop with incorrect second-hop occurs six to ten times more than the opposite (under 3\%), and both remain below 1.2\%. This contrast indicates models rarely achieve correct cultural answers without accurate province identification. Smaller models show even larger variation in gaps, and face difficulties in both  steps.

\begin{figure}[!t]
    \centering
    \begin{subfigure}[t]{0.49\textwidth} 
        \centering
        \includegraphics[width=\textwidth]{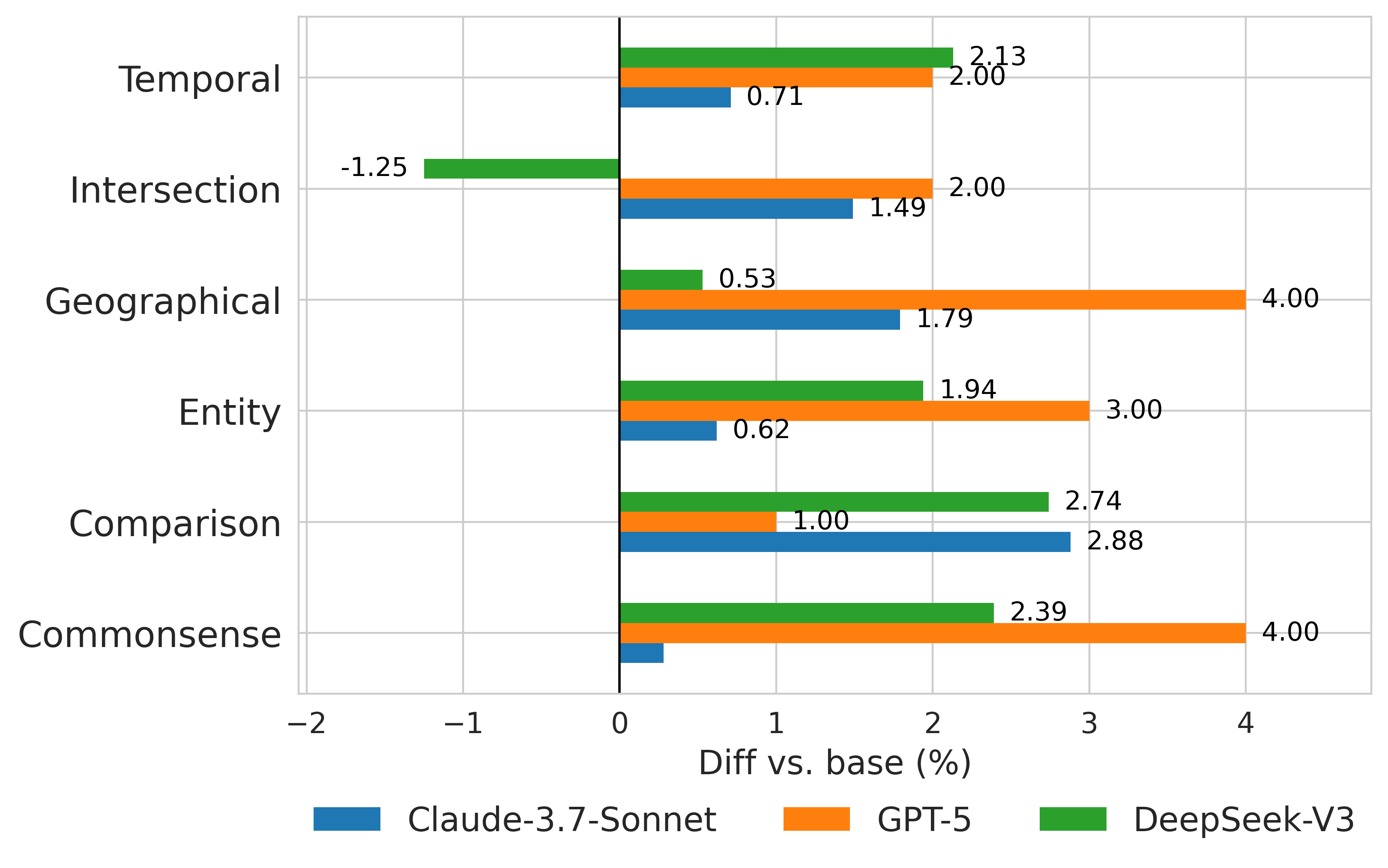}
        \caption{English}
        \label{fig:CoT_improvement_en}
    \end{subfigure}
    \hfill 
    \begin{subfigure}[t]{0.49\textwidth}
        \centering
        \includegraphics[width=\textwidth]{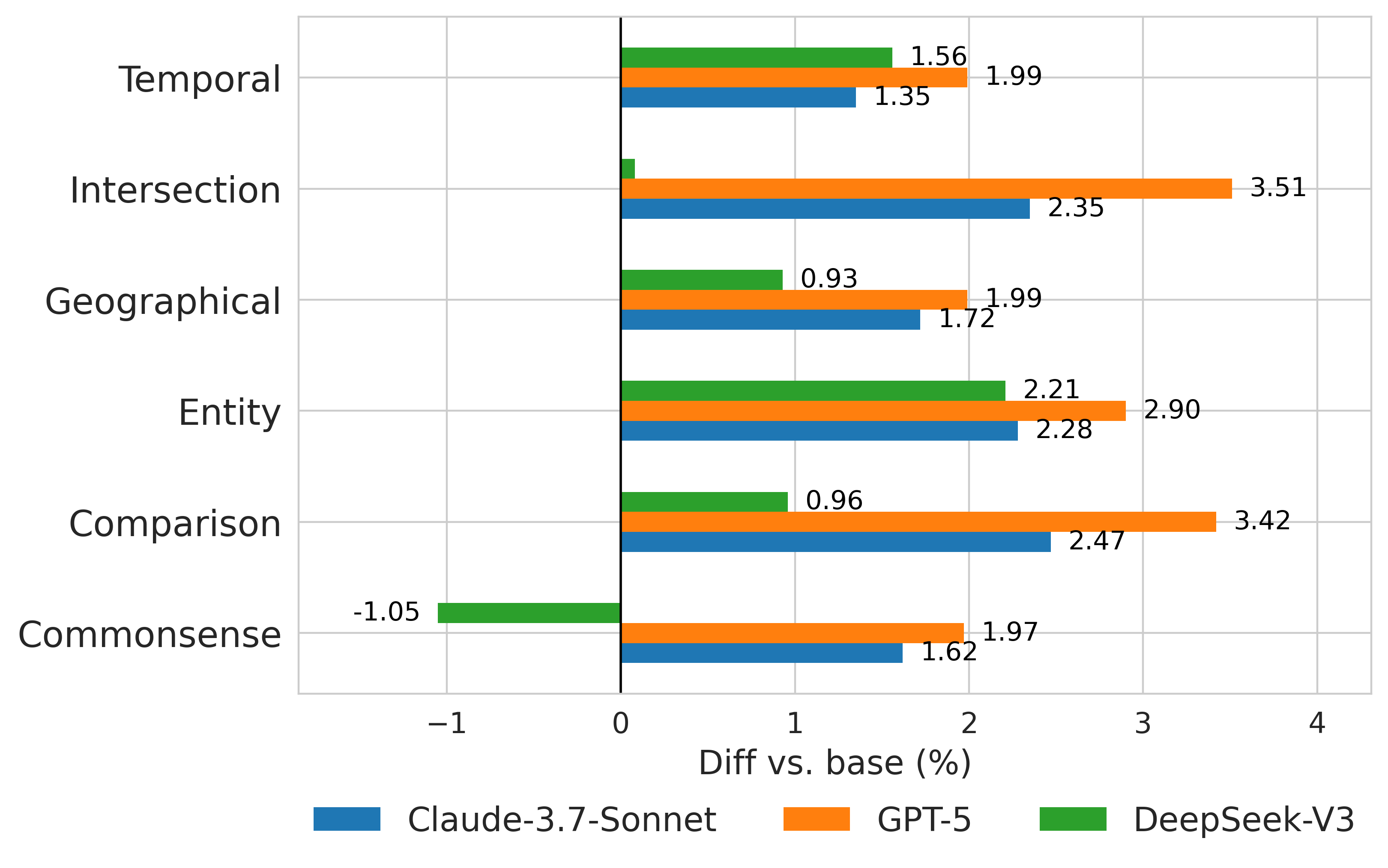}
        \caption{Indonesian}
        \label{fig:CoT_improvement_id}
    \end{subfigure}
    \caption{Improvement (\%) from CoT over Non-CoT prompting across models and question types.}
    \label{fig:CoT_improvement_combined}
\end{figure}

\begin{table*}[!t]
\centering
\tiny
\renewcommand{\arraystretch}{1.15}
\begin{tabular}{@{}
  >{\raggedright\arraybackslash}p{1.5cm}
  >{\raggedright\arraybackslash}p{2.3cm}
  >{\raggedright\arraybackslash}p{2.3cm}
  >{\raggedright\arraybackslash}p{2.3cm}
  >{\raggedright\arraybackslash}p{4.9cm}
@{}}
\toprule
\textbf{Cultural Domain} & \textbf{Question Context} & \textbf{LLMs Answers} & \textbf{Correct Answer} & \textbf{Bias Explanation} \\
\midrule
Food & Aceh casual dining outside home & \textit{Kuah beulangong} (ceremonial curry) & \textit{Sate matang} (grilled meat) & Selected elaborate ceremonial dish over practical everyday food appropriate for casual context \\

Pregnancy and kids & Aceh \textit{mee boh kayee} ceremony & 8th month pregnancy & 3rd month pregnancy & Selected 8th month as approximation to common 7th month traditions rather than regional 3rd month practice \\

Wedding  & West Sumatra proposal tradition & Groom's family gives \textit{uang adat} to bride & Bride's family gives \textit{uang adat} to groom & Applied patriarchal payment direction instead of matrilineal practice where bride's family initiates and pays \\

Fisheries and trade & Livestock distribution & Distribute to relatives and neighbors & Sell at \textit{wosi} market per kg & Applied "traditional equals communal" stereotype, missing local market-based practice \\

Wedding & West Sumatra Koto Gadang style wedding & \textit{Suntiang} headdress & Cloth head covering & Selected widely documented Minangkabau headdress over regional specific practice of mentioned Koto Gadang traditions \\

Pregnancy and kids & Bali placenta burial in house yard & Baby recognizes their house & Baby is always protected & Applied literal practical reasoning over spiritual belief \\

Food & Central Java gudeg Solo taste & Sweet taste & Savory taste & Confused regional variants, selecting Yogyakarta gudeg characteristics over Solo (Central Java) gudeg characteristics \\

Death & North Sumatra post-burial family ritual & Hold prayers for 7 nights at home & Make pilgrimage to scatter flowers on grave & Applied Islamic mourning tradition (7-night prayers), reflecting Indonesia's Muslim majority, over Batak Christian practice of immediate grave visitation with flowers, showing religious framework override\\

\bottomrule
\end{tabular}
\caption{Examples contrasting knowledge prominence versus contextual correctness in model selection across Indonesian cultural domains. All examples show systematic failures where all three models (Claude-3.5-Sonnet, DeepSeek-V3, GPT-5) selected the same incorrect answer.}
\label{tab:knowledge-context}
\end{table*}

\subsection{Chain-of-Thought Results}
To evaluate how in-context reasoning influence LLMs on the ID-MoCQA questions, we tested the three frontier LLMs using Zero-shot Chain-of-Thought (CoT) prompting \citet{Kojima2022CoT} by adding \textbf{\textit{"Let's think step by step"}} to the inputs. Appendix~\ref{app:eval_prompts}) shows the full prompt. CoT results in mixed improvements, with \textit{GPT-5} showing the largest overall gains (averaging $2.67\%$ in English, $2.63\%$ in Indonesian), followed by \textit{Claude-3.7-Sonnet} ($1.97\%$ in Indonesian, $1.30\%$ in English) and \textit{DeepSeek-V3} ($1.41\%$ in English, $0.78\%$ in Indonesian). These improvements suggest that CoT can aid cultural inference tasks, aligning with \citet{romanou2024includeevaluatingmultilinguallanguage}.

Figures~\ref{fig:CoT_improvement_en} and~\ref{fig:CoT_improvement_id} reveal variations in CoT performance across question types, models, and languages. \textit{GPT-5} has the largest and most consistent gains, reaching up to $4.00\%$ on both \textsc{Geographical} and \textsc{Commonsense} in English, and $3.51\%$ on \textsc{Intersection} in Indonesian. However, the negative improvements indicate that CoT can be counterproductive for certain model-task-language combinations, introducing noise or misaligned reasoning. The variation of effectiveness between languages and models suggests that cultural reasoning structures may be represented differently across languages \citep{shi2022language}. Overall, while CoT prompting provides benefits for some models, the inconsistent gains and notable negative cases indicate that cultural reasoning remains challenging and cannot be uniformly solved with zero-shot CoT.

\subsection{Qualitative Analysis}
\label{subsec:qualitative}
We observe that models favor well-documented practices over situationally appropriate ones (Table~\ref{tab:knowledge-context}). When questions explicitly describe casual dining contexts, models select \textit{kuah beulangong} (elaborate ceremonial curry) over \textit{sate matang} (everyday grilled meat). This bias extends to pregnancy ceremonies: models choose 8th-month rituals, likely as approximation to widely documented 7th-month traditions, rather than Aceh's regional 3rd-month \textit{mee boh kayee} ceremony. Papua failures reveal models' "traditional equals communal" stereotypes about indigenous practices. When questions describe pig slaughter in contexts with \textit{bakar batu} stone cooking traditions, models correctly associate \textit{bakar batu} with Papua and successfully identify the province. However, because \textit{bakar batu} is a traditional cultural practice, models then apply "traditional practice equals communal sharing" logic, expecting pigs to be distributed freely to relatives and neighbors. The correct answer is that pigs are sold at \textit{wosi} markets per kilogram, reflecting common practice among locals. Models correctly identify Papua but then misunderstand how cultural practices happen in day-to-day local customs.

\section{Conclusion}
We proposed a framework for expanding single-hop cultural questions into multi-hop questions targeting Indonesian culture. Our resulting multi-hop QA dataset, \textbf{ID-MoCQA}, consists of 15,590 multi-hop questions in Indonesian and English. Our systematic evaluation across ten open-weight and frontier LLMs shows that they struggle with the multi-hop cultural questions. They tend to choose the most well-known cultural information, regardless of whether it is suitable for the specific situation. In the future, we will explore debiasing methods \cite{ko-etal-2020-look,zheng2024large} to mitigate LLMs' preference towards prominent culture. Preference-tuning approaches might also help alleviate LLMs' biases and steer them towards local cultural practices.

\section*{Acknowledgments}
VA is supported by Indonesia Endowment Fund for Education (LPDP), under the Ministry of Finance, Indonesia. XT and NA are supported by the EPSRC [grant number EP/Y009800/1], through funding from Responsible AI UK (KP0016) as a Keystone project. We also acknowledge IT Services at the University of Sheffield for the provision of services for High Performance Computing.

\bibliography{tacl2021}
\bibliographystyle{acl_natbib}

\clearpage
\onecolumn
\appendix
\section{Multi-Hop Question Prompt Guidelines}
\label{app:multi-hop}
\addcontentsline{toc}{section}{Appendix A: Multi-Hop Question Prompt Guidelines}
\subsection{Sample Full Prompt}
\label{app:multi-hopPrompt}
\begin{tcolorbox}[colback=gray!5!white, colframe=black!70!white, title=Sample Full Prompt: Entity]
\scriptsize
\textbf{INSTRUCTIONS}
\textit{Context Conversion:}
\begin{itemize}[nosep]
\item Convert each context'' statement into a culture-related, province-specific question answerable with the provided options''.
\item \textbf{DO NOT} mention the province name in the question.
\item Keep the original topic intact.
\end{itemize}
\textit{Multi-Hop Question Generation:}
\begin{itemize}[nosep]
\item Generate multi-hop questions requiring reasoning to identify the province:
\begin{itemize}
\item \textbf{DO NOT} mention the province, cities, or regencies directly or indirectly --- use only indirect \textbf{CULTURAL} clues.
\item Ensure the correct answer remains consistent with the original options.
\item Avoid repetitive use of similar entities across different questions.
\end{itemize}
\end{itemize}

\textit{Clue Type: Entity}
\begin{itemize}[nosep]
\item Connect the province through a person/thing.
\item Use \textbf{ONLY exact entity names} (people, historical figures, cultural artifacts).
\item Avoid geographical features like lakes, rivers, mountains.
\item \textbf{DO NOT} use compound entities connected with ``and'' --- focus on one clear, specific entity.
\item Ensure the entity is uniquely associated with only this province.
\item \textbf{NEVER} translate, replace, or use synonyms for proper names, historical events, cultural artifacts, or other essential cultural elements.
\end{itemize}
\textit{Examples:}
\begin{itemize}[nosep]
\item ``What musical instrument is commonly featured to wedding ceremonies in the province where Cut Nyak Dhien led resistance against Dutch colonialism?''     
\item ``What traditional dance is performed in the province where former President Susilo Bambang Yudhoyono spent his childhood?''
\item ``What ancient temple complex can be found in the province where coffee variety ‘Toraja’ originates?''
\item ``hat ancient fortress can be visited in the province where pala’ was once worth more than gold to European traders?''
\item ``What traditional harvest celebration performed to express gratitude for blessings and abundance takes place in the province where rendang' was named the world's most delicious food by CNN?''
\end{itemize}
\textit{Language Output:}
\begin{itemize}[nosep]
\item Provide both English and Indonesian versions of the question.
\end{itemize}
\end{tcolorbox}

\begin{table}[h!]
\scriptsize
\centering
\begin{tabular}{|>{\raggedright\arraybackslash}p{7.7cm}|>{\raggedright\arraybackslash}p{7.7cm}|}
\hline
\textbf{Entity} &
\textbf{Geographical} \\
\hline
Connect the province through a person/thing
\textbf{IMPORTANT requirements:}
\begin{itemize}
    \setlength{\itemsep}{0pt}
    \item Use ONLY exact entity names (people, historical figures, cultural artifacts)
    \item Avoid geographical features like lakes, rivers, mountains
    \item Do NOT use compound entities connected with "and" – focus on one clear, specific entity
    \item Ensure the entity is uniquely associated with only this province
    \item NEVER translate, replace, or use synonyms for proper names, historical events, cultural artifacts, or other essential cultural elements
\end{itemize}
&
Connect the province through the location entity
\textbf{IMPORTANT requirements:}
\begin{itemize}
    \setlength{\itemsep}{0pt}
    \item Use ONLY geographical features that are located EXCLUSIVELY in the target province
    \item NEVER use geographical features that cross provincial boundaries (rivers, mountain ranges, watersheds, etc.)
    \item NEVER translate, replace, or use synonyms for proper names, historical events, cultural artifacts, or other essential cultural elements
\end{itemize}
\\
\hline
\textbf{Temporal} &
\textbf{Commonsense Reasoning} \\
\hline
Connect the province through a temporal event (historical or contemporary)
\textbf{IMPORTANT requirements:}
\begin{itemize}
    \setlength{\itemsep}{0pt}
    \item Use significant events with SPECIFIC dates or time periods (historical OR recent)
    \item Temporal references can include, but are not limited to: Ancient \& Pre-Colonial Events, Colonial Period, Independence Era, Modern Historical Events, Natural Disasters, Contemporary Developments, Cultural Milestones, Political Changes, Economic Transformations
    \item Ensure the temporal event is uniquely associated with only this province
    \item Include clear dates or time periods (years, centuries, decades)
    \item NEVER translate, replace, or use synonyms for proper names, historical events, cultural artifacts, or other essential cultural elements
\end{itemize}
&
Formulate a single integrated question that:
\begin{enumerate}
    \setlength{\itemsep}{0pt}
    \item Begins with "If" followed by a concise scenario related to the original context
    \item The scenario should uniquely identify the province through distinctive cultural attributes WITHOUT naming it
    \item Then directly asks about the cultural element from the original context/options
    \item The question should require connecting the scenario (first hop) to the cultural element (second hop)
\end{enumerate}
\textbf{IMPORTANT:}
\begin{itemize}
    \setlength{\itemsep}{0pt}
    \item Keep the scenario part of the question BRIEF and CONCISE (no more than 2 sentences)
    \item The scenario must provide enough cultural context to uniquely identify the province
    \item Use descriptive language instead of naming popular/familiar items of the province
\end{itemize}
\\
\hline
\textbf{Comparison} &
\textbf{Intersection} \\
\hline
Refer the province through comparison
\textbf{IMPORTANT requirements:}
\begin{itemize}
    \setlength{\itemsep}{0pt}
    \item Each comparison MUST identify EXACTLY ONE province
    \item Use diverse formats: Numeric rankings, Range comparisons, Superlatives, Relative metrics
    \item Avoid compound comparisons with "and" – use single, precise comparisons
    \item Use concrete, verifiable metrics (area, number of temples, cultural sites, etc.)
    \item When using population data, ALWAYS specify the year
    \item AVOID using temperature or climate metrics as these can fluctuate seasonally
    \item NEVER translate, replace, or use synonyms for proper names, historical events, cultural artifacts, or other essential cultural elements
\end{itemize}
&
Find the province meeting multiple conditions
\textbf{IMPORTANT requirements:}
\begin{itemize}
    \setlength{\itemsep}{0pt}
    \item The first statement (S1) MUST identify MULTIPLE provinces (at least 3, MAXIMUM 5)
    \item The second statement (S2) must narrow down to EXACTLY ONE province (the target province)
    \item Both conditions must be DISTINCT cultural, geographical, or historical features
    \item Check intersection: Does exactly ONE province match both C1 and C2?
    \item ONLY after verification passes, formulate combined question.
    \item NEVER translate, replace, or use synonyms for proper names, historical events, cultural artifacts, or other essential cultural elements
\end{itemize}
\\
\hline
\end{tabular}
\caption{Prompts of all the clue types.}
\label{tab:clue_type}
\end{table}

\subsection{Clue Types and Structural Templates}
\label{app:reasoning-types}

\textbf{Entity, Geographical, Temporal}, and \textbf{Commonsense} clue type use the templates in Appendix~\ref{app:multi-hopPrompt}, varying only in reasoning specifications and few-shot examples. (\textbf{\textsc{Comparison}, \textsc{Intersection}}) require enhanced prompts with verification procedures due to their factual complexity. Table \ref{tab:clue_type} shows the details.

\paragraph{Comparison} Embedded verification requires confirming comparative claims against empirical data before question finalization:
{\scriptsize
\begin{verbatim}
VERIFY:
- Claim: [exact comparative metric used]
- Data: [provinces with values, showing why the claim identifies exactly one province]
- Data source/year: [specify year for population data or source for other metrics]
- Unique? [YES/NO]
\end{verbatim}
}
\noindent Failed verification triggers iterative claim revision and re-verification until unique identification achieved.

\paragraph{Intersection} Structured step-by-step verification protocol:
{\scriptsize
\begin{verbatim}
S1: [Brief condition description] → [List AT LEAST 3 provinces with minimal proof]
S2: [Brief condition description] → [List provinces with minimal proof]  
Intersection: [Expected single province]
\end{verbatim}
}
\noindent Ensures S1 identifies multiple provinces, S2 narrows to exactly one target province, and intersection produces intended result.

\clearpage

\section{LLM-as-a-Judge Evaluation Criteria}
\label{app:evaluation_prompt}
Our LLM-as-a-judge framework employs eight criteria, each scored 0-2 (16-point maximum), where 2 indicates the highest quality and 0 the lowest.

\subsection{Province Identification and Structural Quality Criteria}
The first four evaluate provincial clue accuracy, conciseness, cultural alignment, and reasoning structure.

\begin{table}[h]
\centering
\scriptsize
\begin{tabular}{|p{4cm}|c|p{9.5cm}|}
\hline
\textbf{Criterion} & \textbf{Score} & \textbf{Description} \\
\hline
\textbf{Provincial Specificity \& Factual Accuracy} & 2 & Clues uniquely identify one province with accurate information \\
\cline{2-3}
 & 1 & Clues apply to 2-3 provinces with minor ambiguity, no factual errors \\
\cline{2-3}
 & 0 & Clues contain factual errors or hallucinations \\
\hline
\textbf{Redundancy of Province Clues} & 2 & Concise, non-repetitive cultural or geographic clues \\
\cline{2-3}
 & 1 & Minor redundancy present, core clue remains meaningful \\
\cline{2-3}
 & 0 & Excessive repetition reduces reasoning complexity \\
\hline
\textbf{Reference Alignment \& Rephrasing Quality} & 2 & Accurately reflects original context with different wording \\
\cline{2-3}
 & 1 & Minor deviations but maintains essential cultural elements \\
\cline{2-3}
 & 0 & Significantly alters or misrepresents cultural information \\
\hline
\textbf{Multi-hop Structure} & 2 & Clear two-step reasoning: province identification, then cultural question \\
\cline{2-3}
 & 1 & Attempts two-step process but partially combines steps \\
\cline{2-3}
 & 0 & Fails to create proper two-step structure \\
\hline
\end{tabular}
\caption{Province identification and structural quality criteria.}
\label{tab:evaluation_criteria_part1}
\end{table}

\vspace{-12pt}

\subsection{Answer Quality and Presentation Criteria}
Four criteria assess reasoning necessity, answer alignment, clarity, and language quality.

\begin{table}[h]
\centering
\scriptsize
\begin{tabular}{|p{4cm}|c|p{9.5cm}|}
\hline
\textbf{Criterion} & \textbf{Score} & \textbf{Description} \\
\hline
\textbf{Answer Discrimination} & 2 & Requires both reasoning steps for correct answer \\
\cline{2-3}
 & 1 & Can narrow to 2 options after one step \\
\cline{2-3}
 & 0 & Correct answer identifiable without province identification \\
\hline
\textbf{Answer Quality} & 2 & Perfect alignment with original options and reasoning process \\
\cline{2-3}
 & 1 & Minor inconsistencies but remains functionally appropriate \\
\cline{2-3}
 & 0 & Significantly alters context, making options inappropriate \\
\hline
\textbf{Question Clarity} & 2 & Clearly articulated and easily understandable \\
\cline{2-3}
 & 1 & Minor clarity issues but remains interpretable \\
\cline{2-3}
 & 0 & Confusing phrasing or disorganised structure \\
\hline
\textbf{Language Quality} & 2 & Grammatically correct with natural phrasing \\
\cline{2-3}
 & 1 & Minor grammar issues but understandable \\
\cline{2-3}
 & 0 & Significant grammar problems hindering understanding \\
\hline
\end{tabular}
\caption{Answer quality and presentation criteria. \textbf{Answer Discrimination} verifies both reasoning steps required (prevents shortcuts). \textbf{Answer Quality} ensures multi-hop-option compatibility and reasoning structure alignment. \textbf{Question Clarity} assesses comprehension, coherence, and organization. \textbf{Language Quality} evaluates grammar and naturalness (both languages) while preserving Indonesian cultural terms.}
\label{tab:evaluation_criteria_part2}
\end{table}

\vspace{-8pt}

\section{Dataset Verification Pipeline: Prompt Engineering}
\label{app:verification_prompts}
\subsection{Phase 1: Issue Detection}
Identifies and corrects option text copying and incorrect province name usage as location references.
{\scriptsize
\begin{verbatim}
1. OPTION TEXT COPYING:
   - Does the MHQA contain phrases copied from the options? Mark as true/false.
2. PROVINCE NAME USAGE:
   - Is the province name used specifically as a location in the MHQA?
   - Cultural terms (e.g., 'Rumoh Aceh') must NOT be marked.
   - Only flag location usage (e.g., 'in Aceh province', 'from Bali').
REVISION GUIDELINES:
- If copying detected: Reword to avoid option text
- If province location detected: Use indirect references
\end{verbatim}
}

Cultural terms (e.g., 'Rumoh Aceh') versus location references (e.g., 'in Aceh province') distinction maintains authenticity while eliminating geographic shortcuts.

\subsection{Phase 2: Quality Assessment}
Evaluates multi-hop structure integrity. CRITICAL: Prevent reintroduction of location-based province references.
{\scriptsize
\begin{verbatim}
CRITICAL RULE: DO NOT use province names as locations in revisions
EVALUATE:
- Proper multi-hop question? (identify province → cultural question)
- Clear sequential reasoning?
- Correct grammar for the language used?
IMPROVE:
1. Two-step structure: province identification → cultural question
2. Use indirect province references only
3. Preserve cultural terms exactly ('Rumoh Aceh', 'Soto Aceh')
4. Fix grammar and maintain natural flow
DECISION:
- Minor fixes needed: REVISE
- Major restructuring needed: mark "[NEEDS MAJOR REVISION]"
\end{verbatim}
}

\section{Manual Evaluation Guideline}
\label{app:annotation-guide}

Three native Indonesian graduate students evaluate ID-MoCQA question quality and difficulty through (1) linguistic naturalness assessment, (2) multi-hop question answering, and (3) cognitive difficulty rating. External sources and AI assistance are prohibited. Annotators receive: \textbf{Context} (IndoCulture premise), \textbf{ID-MoCQA} bilingual questions, and \textbf{Options} (three choices A/B/C in both languages).

\subsection{Task 1: Naturalness}
Rate linguistic and cultural naturalness on a 3-point scale:
\begin{itemize}[nosep]
    \item \textbf{Natural}: Fluent, grammatically correct, culturally accurate, and sounds authentic. 
    
    \item \textbf{Acceptable}: Understandable with minor issues (slight awkwardness, minor grammar errors, somewhat unnatural phrasing). 
    
    \item \textbf{Unnatural}: Major grammatical errors, very awkward phrasing, culturally incorrect references, or incomprehensible.     
\end{itemize}

\subsection{Task 2: Multi-hop Question Answering}
\vspace{-0.3em}
\textbf{Objective:} Answer each question through a two-step reasoning process that connects cultural context to the correct answer.

Annotators perform the following steps:
\begin{enumerate}[nosep]
    \item \textbf{Identify the target province:} Use the provided cultural clues to determine which Indonesian province is being referenced.
    \item \textbf{Select the correct answer:} Choose one option (A, B, or C) that correctly answers the question based on the identified province.
\end{enumerate}
\vspace{-8pt}

\subsection{Task 3: Difficulty Assessment}
\vspace{-0.3em}
\textbf{Objective:} Assess the cognitive difficulty required to answer each question.

Annotators assign one difficulty label to each question based on the reasoning complexity and the rarity of cultural knowledge required:
\begin{itemize}[nosep]
    \item \textbf{Easy:} The question involves widely known cultural facts and can be answered with minimal reasoning or common knowledge.
    \item \textbf{Moderate:} The question requires moderate reasoning or specific cultural knowledge that may not be universally familiar.
    \item \textbf{Hard:} The question requires specialized regional cultural knowledge, uncommon facts, or complex multi-hop reasoning to derive the correct answer.
\end{itemize}

\section{Semantic Analysis}
\label{app:semantic}
\vspace{-4pt}

We extracted lexical and semantic features from all English questions using GPT-4o-mini (temperature=0). English questions were analyzed for two reasons: (1) current LLMs provide more reliable part-of-speech tagging and named entity recognition for English; (2) Indonesian cultural terms (traditional items, ceremonies, place names) remain identical across both language versions, ensuring cultural authenticity is captured regardless of analysis language. The bilingual dataset's parallel structure ensures lexical patterns in English reflect the same cultural content as Indonesian versions.
\vspace{-4pt}
\begin{itemize}[nosep]
    \item \textbf{Lexical features}: Adjectives, nouns, and verbs.
    \item \textbf{Named entities}: Person names and location names.
    \item \textbf{Temporal expressions}: Time references (historical periods, dates, contemporary events).
    \item \textbf{Indonesian cultural terms}: Culture-specific words preserved in original language.
    \item \textbf{Word count}: Total words per question.
\end{itemize}

\begin{tcolorbox}[colback=gray!5!white, colframe=black!70!white, title=Prompt Format: Semantic Analysis]
\scriptsize
\begin{verbatim}
Analyze this Indonesian culture question.
Question: "{question}"
Extract (return ONLY valid JSON)
\end{verbatim}
\end{tcolorbox}

\section{Model Evaluation Prompts}
\label{app:eval_prompts}
\subsection{Evaluation Configuration}
Models are evaluated with temperature=0 (except GPT-5: default=1.0, no temperature support) via APIs in zero-shot settings. The Zero-shot prompt asks for structured output (province + answer) without explanations. The CoT prompt requests step-by-step reasoning before structured output.

\subsection{Zero-shot Evaluation Prompt Structure}
\begin{tcolorbox}[colback=gray!5!white, colframe=black!70!white, title=Prompt Format: Indonesian]
\scriptsize
\begin{verbatim}
Jawab HANYA dengan format: PROVINSI: [nama provinsi] JAWABAN: [huruf]. Tanpa penjelasan.
Pertanyaan: {pertanyaan}
Pilihan: A. {opsi_a} B. {opsi_b} C. {opsi_c}

\end{verbatim}
\end{tcolorbox}
 
\begin{tcolorbox}[colback=gray!5!white, colframe=black!70!white, title=Prompt Format: English]
\scriptsize
\begin{verbatim}
ONLY respond with format: PROVINCE: [province name] ANSWER: [letter]. No explanations.
Question: {question}
Options: A. {option_a} B. {option_b} C. {option_c}

\end{verbatim}
\end{tcolorbox}

\subsection{Zero-shot Chain of Thought (CoT) Evaluation Prompt Structure}
\begin{tcolorbox}[colback=gray!5!white, colframe=black!70!white, title=CoT Prompt Format: Indonesian]
\scriptsize
\begin{verbatim}
Pertanyaan: {pertanyaan}
Pilihan: A. {opsi_a} B. {opsi_b} C. {opsi_c}

Mari berpikir langkah demi langkah untuk menjawab pertanyaan ini. 
Pertama, prediksi provinsi Indonesia yang paling terkait dengan pertanyaan ini.
Kemudian, analisis pertanyaan dan pilihan untuk menentukan jawaban yang benar.
Analisis: [Jelaskan proses pemikiran langkah demi langkah]
Setelah memberikan penjelasan, akhiri jawaban dengan format:
Provinsi: [nama provinsi]
Jawaban: [A/B/C]
\end{verbatim}
\end{tcolorbox}
 
\begin{tcolorbox}[colback=gray!5!white, colframe=black!70!white, title=CoT Prompt Format: English]
\scriptsize
\begin{verbatim}
Question: {question}
Options: A. {option_a} B. {option_b} C. {option_c}

Let's think step by step to answer this question.
First, predict which Indonesian province this question is most associated with.
Then, analyse the question and options to determine the correct answer.
Analysis: [Explain your step-by-step thinking process]
After providing your explanation, end your answer with this format:
Province: [province name]
Answer: [A/B/C]
\end{verbatim}
\end{tcolorbox}

\section{Additional Analysis}

\subsection{Human-Human Disagreement Examples}
\label{app:human_disagreements} 
During the multi-annotator validation process \hyperref[subsec:LLM-as-a-Judge]{§4.2}, we observed several patterns in cases where human annotators disagreed on question quality. Table~\ref{tab:disagreement-examples} presents two examples illustrating the types of ambiguities that led to disagreement and how they were resolved through majority vote. These cases demonstrate that disagreements often stem from legitimate differences in judgment criteria rather than annotation errors, particularly regarding: (1) the level of temporal and statistical specificity required for comparative claims, and (2) whether certain question structures might reveal answers.

\begin{table}[h]
\centering
\scriptsize
\begin{tabular}{p{0.15\textwidth}p{0.75\textwidth}}
\toprule
\textbf{Component} & \textbf{Details} \\
\midrule
\multicolumn{2}{l}{\textbf{Example 1: Temporal Specificity in Comparative Claims}} \\
\midrule
\textbf{Question} & After grandfather Ali was buried, what activity did Ali perform for 7 or 10 days 
in the region with the highest percentage of Muslim population in Indonesia? \\
\addlinespace
\textbf{Options} & A. Ali holds \textit{tahlilan} at home \newline 
B. Ali goes to work immediately \newline 
C. Ali recites \textit{yasin} prayers at the gravesite for 7 or 10 nights \\
\addlinespace
\textbf{Correct Answer} & C \\
\addlinespace
\textbf{Annotator 1} & \textit{Minor} \\
\addlinespace
\textbf{Annotator 2} & \textit{Moderate} \\
\addlinespace
\textbf{Annotator 3} & \textit{Moderate} -- Without explicit official statistic data (and specificity mentioned year), comparative claim lacks verifiability. Though Aceh's Sharia law provides cultural signal, statistical 
claim needs citation \\
\addlinespace
\textbf{Final Decision} & \textbf{Moderate} (majority vote: 2/3) \\
\addlinespace
\textbf{Rationale} & Comparative statistical claims should reference specific, verifiable data to 
avoid ambiguity \\
\midrule
\multicolumn{2}{l}{\textbf{Example 2: Answer Discrimination and Structural Issues}} \\
\midrule
\textbf{Question} & If Budi lives in a region that formally implements Islamic Sharia law with 
qanun (special regional regulations) governing public behavior and has Wilayatul Hisbah 
as Sharia police, what traditional beverage would he buy as a souvenir for his father? \\
\addlinespace
\textbf{Options} & A. Budi buys \textit{Emping Melinjo} \newline 
B. Budi buys \textit{kopi Gayo} \newline 
C. Budi buys \textit{Kue Bhoi} \\
\addlinespace
\textbf{Correct Answer} & B \\
\addlinespace
\textbf{Annotator 1} & \textit{Minor} \\
\addlinespace
\textbf{Annotator 2} & \textit{Moderate} \\
\addlinespace
\textbf{Annotator 3} & \textit{Minor} -- Though only one option is a beverage (option B), question retains complexity by testing both province identification (Sharia law) and cultural knowledge 
(traditional beverage) \\
\addlinespace
\textbf{Final Decision} & \textbf{Minor} (majority vote: 2/3) \\
\addlinespace
\textbf{Rationale} & Question retains sufficient cultural reasoning complexity despite having only 
one beverage option, as it requires identifying the region through legal/cultural markers 
before selecting the appropriate souvenir \\
\bottomrule
\end{tabular}
\caption{Examples of human annotator disagreements and the resolution through majority vote. }
\label{tab:disagreement-examples}
\end{table}

\subsection{Distribution of Culture-Specific Terms Across Provinces}
As discussed in \hyperref[subsec:final-dataset]{§4.6}, provincial distribution shows variation in cultural-linguistic specificity. East Nusa Tenggara (226 questions) has the highest density of culture-specific terms preserved in local language (0.92 terms per question), suggesting its cultural practices rely heavily on local terminology. In contrast, West Sumatra, despite having the most questions (1,072), shows lower cultural term density (0.65 terms per question), indicating its cultural practices may be more well known or describable with general Indonesian vocabulary. This pattern appears across other provinces: East Java (469 questions, 0.81 terms per question), Central Java (653 questions, 0.81 terms per question), and Aceh (808 questions, 0.76 terms per question) maintain higher cultural-linguistic specificity than the three most-represented provinces (West Sumatra, Papua, and North Sumatra), which average 0.68 terms per question.

\subsection{Error patterns of human performance}
Analysis of the human baseline reported in \hyperref[subsec:human-performance]{§6.1} reveals systematic patterns across topics and provinces. Food-related questions accounted for 15-16\% of all errors, followed by Wedding (15-16\%) and Art-related questions (12-16\%). Province-level error rates varied substantially: West Sumatra showed error rates of 32-42\% across participants, followed by South Sulawesi (31-45\%) and Papua (29-41\%). In contrast, West Java showed the lowest error rates at 9-15\%, followed by Bali (13-17\%) and Central Java (16-27\%). Notably, provinces with the lowest error rates are among Indonesia's most well-known regions both domestically and internationally.

\subsection{Performance Consistency Across Clue Types}
Beyond the overall accuracy patterns shown in Table~\ref{tab:accuracy_MHQA}, performance balance across clue types varies more by individual model characteristics than by scale alone. \textit{GPT-5} shows the narrowest performance range at approximately 2.8 percentage points in English and 2.7 points in Indonesian between its best and worst types, maintaining consistency across all six reasoning categories. \textit{Claude-3.7-Sonnet} demonstrates similarly tight ranges of 1.9 points in English and 2.6 points in Indonesian, indicating highly balanced capabilities. \textit{Llama3.3-70B-IT} shows moderate ranges around 3.2 points in both languages. However, some smaller models display comparable stability: \textit{SeaLLM-7B} shows ranges of approximately 1.7 points in English and 2.3 points in Indonesian, \textit{Qwen2.5-7B} shows approximately 4.7 points in English and 3.3 points in Indonesian, and \textit{Merak-7B} demonstrates ranges around 3.9 points in English and 4.2 points in Indonesian. These patterns indicate that frontier models develop more balanced reasoning capabilities, while some smaller models show greater variability.

\subsection{Multi-Hop Reasoning Error Patterns}
Figure~\ref{tab:firsthop_secondhop_breakdown} reveals that performance gaps widen across model scales. \textit{Llama3.3-70B-IT}, \textit{Qwen2.5-72B-IT}, and \textit{Gemma2-27B-IT} show 28-30 point gaps with incorrect first-hop but correct second-hop reaching up to 8.3\%. Smaller models show even larger variation in gaps (16-35 points across both languages), with \textit{Llama3.1-8B} at 35 points in English and \textit{Qwen2.5-7B} showing incorrect first-hop but correct second-hop at 23.3\% and both incorrect at 22.1\%. \textit{Merak-7B} shows approximately 30-point gaps with both incorrect reaching 17-18\% and first-hop accuracy around 66-67\% despite Indonesian training. \textit{SeaLLM-7B} demonstrates smaller gaps (16-20 points) but lower overall first-hop accuracy (51-53\%) and higher both-incorrect rates (18-20\%). These patterns indicate smaller models face difficulties at both reasoning steps, with elevated reverse error rates suggesting occasional reliance on alternative reasoning pathways that do not depend on accurate province identification.

Cross-linguistic comparison reveals that language effects vary by model category. Frontier models show minimal changes (0.6-0.8 point decreases in first-hop correct but second-hop incorrect), while \textit{Llama3.3-70B-IT} increases both-correct by 3.6 points in Indonesian, demonstrating target-language presentation specifically reduces cultural reasoning errors. In contrast, \textit{Merak-7B}'s both-correct declines 1.6 points despite language-specific training. Overall, first-hop to both-correct gaps widen as model performance decreases (frontier: 18-23 points; 70B models: 28-30 points; smaller models: 16-35 points), suggesting that weaker models accumulate errors across the reasoning chain rather than failing at specific steps.

\subsection{Qualitative Error Analysis}
We examined the failure cases of GPT-5, Claude-3.7-Sonnet, and DeepSeek-V3 models reported in \hyperref[subsec:qualitative]{§6.4}. In most of the cases, all three models choose the same incorrect answer, indicating shared systematic biases. Topics like death ceremonies, traditional games, and art forms exhibit substantially higher same-wrong-answer rates than daily activities.

Models consistently demonstrate strong province identification (averaging 96.5\%) but struggle with cultural reasoning within correctly identified contexts. West Sumatra exemplifies this pattern: models recognize matrilineal cultural markers yet systematically apply patriarchal logic. In the \textit{bajapuik} wedding tradition, the bride's family pays \textit{uang japuik} to the groom's family, reflecting matrilineal practice. However, models incorrectly expect the groom's family to pay, following widespread patriarchal dowry patterns. Central Java demonstrates unique identification challenges (89\% accuracy), driven by cultural similarity rather than geographic proximity. Models frequently confuse Central Java with other Javanese provinces (West Java, East Java, Yogyakarta) that share gamelan music, batik arts, and court traditions. In contrast, Papua, though geographically isolated, demonstrates strong identification. Models struggle to distinguish provinces sharing similar cultural features. Javanese regions exemplify this: despite individual prominence, models confuse those sharing gamelan music, batik arts, and court traditions.

\end{document}